\ifdefined\XeTeXversion\else\pdfoutput=1\fi
\documentclass[10pt]{article}
\usepackage[T1]{fontenc}
\usepackage[letterpaper,textwidth=150mm,top=25mm,bottom=24mm,headheight=14pt,headsep=10pt,footskip=12mm]{geometry}
\usepackage{graphicx,xcolor,amsmath,amssymb,booktabs,tabularx,longtable,array,multirow}
\usepackage{newtxtext,newtxmath}
\usepackage{enumitem}
\usepackage[font=small,labelfont=bf,skip=6pt]{caption}
\usepackage{float}
\usepackage{placeins}
\usepackage{fancyhdr}
\usepackage{titletoc}
\usepackage{xurl}
\usepackage[colorlinks=true,linkcolor=PKAccent,citecolor=PKAccent,urlcolor=PKAccent]{hyperref}
\usepackage{tikz}
\usetikzlibrary{arrows.meta,positioning,calc,fit,backgrounds}

\definecolor{PKInk}{HTML}{111F2B}
\definecolor{PKAccent}{HTML}{163A5F}
\definecolor{PKBlue}{HTML}{2B6CB0}
\definecolor{PKTeal}{HTML}{0F766E}
\definecolor{PKCoral}{HTML}{C2410C}
\definecolor{PKAmber}{HTML}{B45309}
\definecolor{PKSlate}{HTML}{64748B}
\definecolor{PKMuted}{HTML}{505D6D}
\definecolor{PKRule}{HTML}{DCE3EA}
\definecolor{PKPaper}{HTML}{F8FAFC}
\definecolor{PKTint}{HTML}{EEF4F8}

\setlist[itemize]{nosep,leftmargin=*}
\makeatletter
\newcommand{\ds@subtitle}{}
\newcommand{\ds@affiliation}{}
\newcommand{\ds@runninghead}{DeltaScience Technical Report}
\newcommand{\dssubtitle}[1]{\renewcommand{\ds@subtitle}{#1}}
\newcommand{\dsaffiliation}[1]{\renewcommand{\ds@affiliation}{#1}}
\newcommand{\dsrunninghead}[1]{\renewcommand{\ds@runninghead}{#1}}
\newif\ifdslogoleft
\newcommand{\dslogo}{\includegraphics[width=42mm]{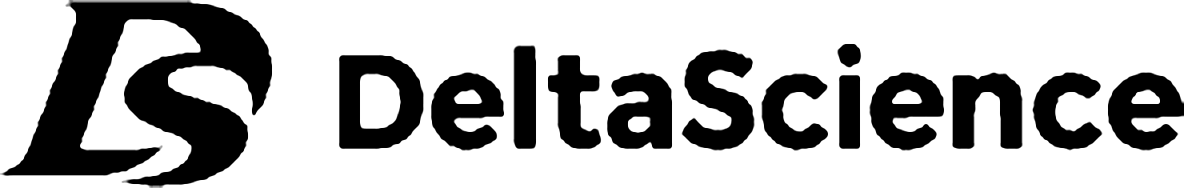}}
\@namedef{ds@brand@deepseek}{\includegraphics[height=1.15em]{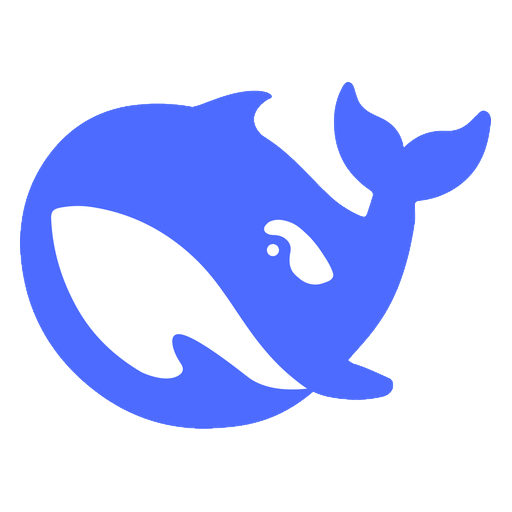}}
\@namedef{ds@brand@doubao}{\includegraphics[height=1.15em]{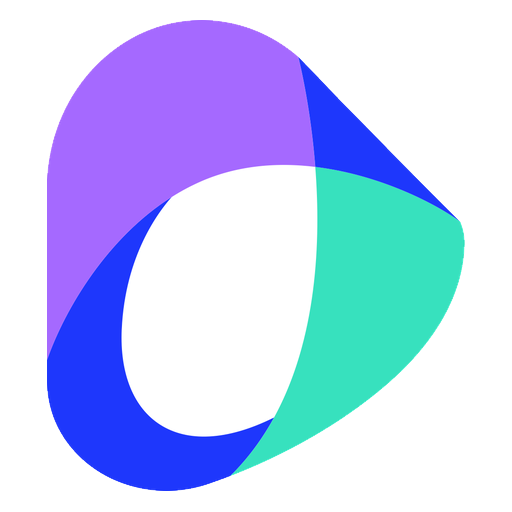}}
\@namedef{ds@brand@glm}{\includegraphics[height=1.15em]{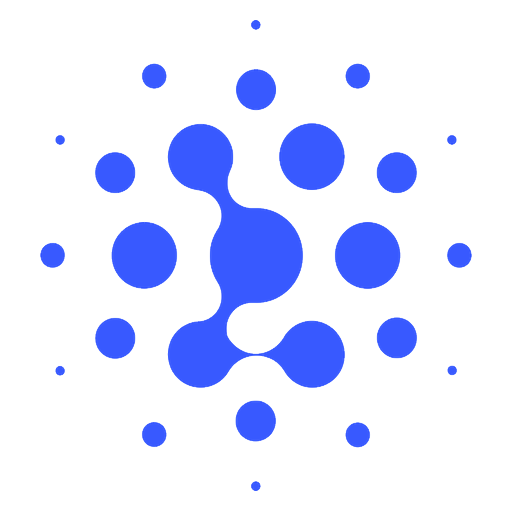}}
\@namedef{ds@brand@gpt}{\includegraphics[height=1.15em]{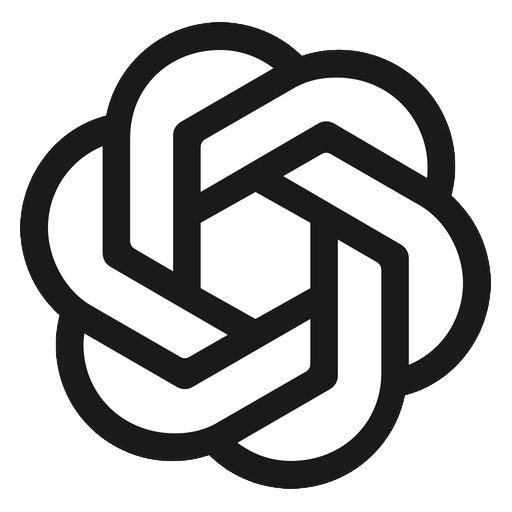}}
\@namedef{ds@brand@hy4}{\includegraphics[height=1.15em]{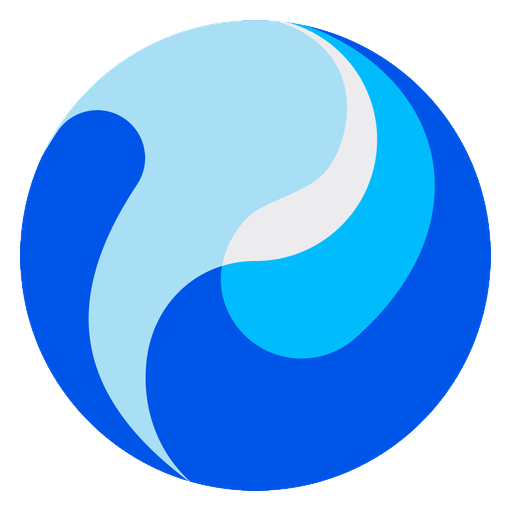}}
\@namedef{ds@brand@kimi}{\includegraphics[height=1.15em]{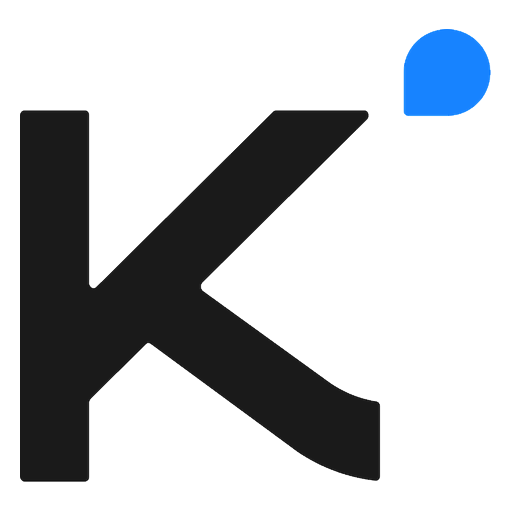}}
\@namedef{ds@brand@mimo}{\includegraphics[height=1.15em]{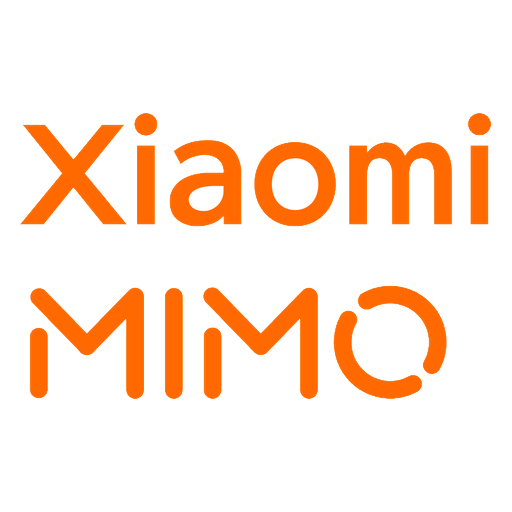}}
\@namedef{ds@brand@qwen}{\includegraphics[height=1.15em]{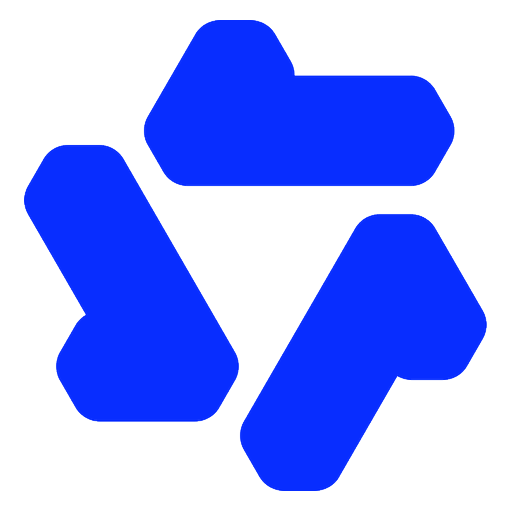}}
\newcommand{\modellogo}[1]{\raisebox{-0.22em}{\@nameuse{ds@brand@#1}}\hspace{0.4em}}
\author{}
\date{}

\fancypagestyle{plain}{%
  \fancyhf{}%
  \fancyfoot[C]{\footnotesize\color{PKMuted}\thepage}%
}

\renewcommand{\maketitle}{%
  \thispagestyle{plain}%
  \begingroup
  \setlength{\parskip}{0pt}%
  \noindent
  \ifdslogoleft
    \raisebox{-0.5\height}{\dslogo}\hfill
    \parbox[c]{0.65\linewidth}{\raggedleft\small\color{PKMuted}\@date}%
  \else
    \parbox[c]{0.65\linewidth}{\raggedright\small\color{PKMuted}\@date}\hfill
    \raisebox{-0.5\height}{\dslogo}%
  \fi\par
  \vspace{8pt}%
  {\color{PKRule}\hrule height 0.6pt}\vspace{16pt}%
  {\raggedright\LARGE\bfseries\color{PKInk}\@title\par}%
  \ifx\ds@subtitle\@empty\else
    \vspace{5pt}%
    {\raggedright\large\color{PKMuted}\ds@subtitle\par}%
  \fi
  \ifx\@author\@empty\else
    \vspace{12pt}%
    {\raggedright\normalsize\color{PKInk}\@author\par}%
  \fi
  \ifx\ds@affiliation\@empty\else
    \vspace{4pt}%
    {\raggedright\small\color{PKMuted}\ds@affiliation\par}%
  \fi
  \vspace{6pt}%
  {\raggedright\footnotesize\color{PKMuted}\textbf{Platform:} \url{https://deltascience.bohrium.com} \quad\textbar\quad \textbf{Contact:} \href{mailto:deltascience@aisi.ac.cn}{\texttt{deltascience@aisi.ac.cn}}\\[2pt]\textbf{GitHub:} \url{https://github.com/deepmodeling/DeltaScience}\par}%
  \endgroup
  \@thanks
  \par\vspace{12pt}%
}

\renewenvironment{abstract}{%
  \begin{list}{}{%
    \setlength{\leftmargin}{8mm}%
    \setlength{\rightmargin}{8mm}%
    \setlength{\topsep}{0pt}%
    \setlength{\parsep}{4pt}%
    \setlength{\itemindent}{0pt}}%
  \item[]\small
  {\centering\bfseries\color{PKInk}\abstractname\par}\nobreak\vspace{5pt}%
}{\end{list}\par\medskip}
\makeatother

\tikzset{
  pkbox/.style={draw=PKRule,fill=PKPaper,rounded corners=2pt,align=center,
    font=\small,text=PKInk,minimum height=12mm,inner sep=3pt},
  pkarrow/.style={-{Stealth[length=2mm]},draw=PKMuted,line width=0.5pt},
  pknote/.style={font=\footnotesize,text=PKMuted}}

\newcommand{\DSNumTasks}{65}
\newcommand{\DSNumCombos}{11}
\newcommand{\DSNumRuns}{3,575}
\newcommand{\DSNumGroups}{7}
\newcommand{\DSNumSubfields}{35}
\newcommand{\DSCrossCount}{29}
\newcommand{\DSCrossPct}{44.6\%}
\newcommand{\DSTopScore}{61.3}
\newcommand{\DSTopCI}{[51.3, 70.6]}
\newcommand{\DSSecondScore}{56.0}
\newcommand{\DSThirdScore}{44.0}
\newcommand{\DSFourthScore}{40.5}
\newcommand{\DSFifthScore}{38.3}
\newcommand{\DSNeverFullPct}{61.1\%}
\newcommand{\DSAllFiveFullPct}{13.8\%}
\newcommand{\DSMedWithinSD}{4.0}
\newcommand{\DSVarTask}{52.3\%}
\newcommand{\DSVarCombo}{15.9\%}
\newcommand{\DSVarResid}{31.8\%}
\newcommand{\DSHarnessGap}{$+11.0$}
\newcommand{\DSHarnessCI}{$[+6.7, +15.4]$}
\newcommand{\DSHarnessWL}{42/10}
\newcommand{\DSInterCorr}{0.60}
\newcommand{\DSInterCorrRange}{0.26--0.86}
\newcommand{\DSOracleMean}{69.6}
\newcommand{\DSOracleGain}{$+8.4$}
\newcommand{\DSSolvedAny}{47}
\newcommand{\DSUnsolvedAll}{6}
\newcommand{\DSNonTopBeats}{12}
\newcommand{\DSRhoTime}{-0.56}
\newcommand{\DSRhoOtok}{-0.09}
\newcommand{\DSPairSep}{45}
\newcommand{\DSPairTotal}{55}
\newcommand{\DSTaskQOne}{15.7}
\newcommand{\DSTaskMed}{34.5}
\newcommand{\DSTaskQThree}{50.0}
\newcommand{\DSTaskMax}{98.2}

\newcommand{\DSTopPassFive}{60.0\%}
\newcommand{\DSTopPowFive}{49.2\%}
\newcommand{\DSTopOutcomeTagEN}{Target selectivity}
\newcommand{\DSTopOutcomeRate}{60.0\%}
\newcommand{\DSGlmOnTime}{35.6}
\newcommand{\DSGlmLoss}{11.7}
\newcommand{\DSDoubaoOnTime}{31.6}
\newcommand{\DSDoubaoLoss}{11.8}
\newcommand{\DSDsOnTime}{35.3}
\newcommand{\DSHyDshOnTime}{40.8}
\newcommand{\DSHyDshLoss}{2.5}
\newcommand{\DSHyCbOnTime}{32.2}
\newcommand{\DSHyCbLoss}{4.9}
\newcommand{\DSTimeLimitMin}{1}
\newcommand{\DSTimeLimitMed}{8}
\newcommand{\DSTimeLimitMean}{8.4}
\newcommand{\DSTimeLimitMax}{48}
\newcommand{\DSTimeLimitGeEightN}{42}
\newcommand{\DSTimeLimitGeEightPct}{64.6\%}

\dsrunninghead{DeltaScience Technical Report}
\title{DeltaScience: Benchmarking AI Agents on Real Scientific Problems}
\dssubtitle{An Open Evaluation Platform for Large Models and AI Agents in Real Scientific Scenarios}
\author{\textbf{DeltaScience Team}\textsuperscript{*}}
\dsaffiliation{%
AI for Science Institute, Beijing\\
DP Technology\\
Peking University\\
Zhejiang University\\
University of Tokyo\\
University of Cambridge\\
Institute of Physics, Chinese Academy of Sciences\\
Beihang University\\
Tongji University%
}
\date{November 2026\quad Technical Report v1.0}

\begin{document}
\maketitle
{\renewcommand{\thefootnote}{\fnsymbol{footnote}}\footnotetext[1]{Full list of contributors and functional roles appears in the Contributors section at the end of this report.}}

\begin{abstract}
AI agents are moving from static scientific recall toward autonomous execution of real research workflows, yet most existing benchmarks evaluate isolated QA or code snippets and cannot determine whether an agent's calculation, design, or analysis passes scientific acceptance for downstream use. We introduce \textbf{DeltaScience}, \textbf{an open evaluation platform for large models and AI agents in real scientific scenarios}, dedicated to \textbf{connecting frontier scientific discovery with real-world industrial R\&D, and testing AI scientific capability on problems that truly matter}, ensuring every capability gain reflects a retestable scientific increment ($\delta$) and embodying the continuous evolution of both the benchmark and AI scientific capability. DeltaScience adopts a \textbf{three-layer decoupled governance architecture}: the non-profit DeltaScience Benchmark independently sets scientific acceptance standards, governs domain-expert admission, and publishes versioned releases; Playground (including its year-round authoring competition) operates open community task submissions, batch review, and contributor incentives; and Bohrium supplies isolated sandboxes and compute infrastructure, separating evaluation standards from execution.

DeltaScience v1.0 comprises \textbf{\DSNumTasks{} private real-world tasks} spanning \DSNumGroups{} discipline groups, \DSNumSubfields{} subfields, and 4 mutually exclusive task types, with \DSCrossCount{} tasks (\DSCrossPct{}) crossing multiple discipline groups. Eleven versioned \textbf{Model + Harness} configurations execute five independent isolated trials per task (\DSNumRuns{} runs total). Deep empirical analysis yields six primary findings: (1) GPT-6.2 Astra + Codex leads at \textbf{\DSTopScore{}} (95\% CI \DSTopCI{}), ahead of GPT-6.2 Sol + Codex (\DSSecondScore{}) and Tier 2 entrants Qwen4-Max + Qoder (\DSThirdScore{}), Kimi K3.1 Pro + Kimi Code (\DSFourthScore{}), and HY4 + DSH (\DSFifthScore{}), with \DSPairSep{} of \DSPairTotal{} configuration pairs statistically separated; (2) multi-run reliability remains a major bottleneck---under per-task scientific pass criteria, even the \#1 system drops from pass@5 = \DSTopPassFive{} to pass\textsuperscript{5} = \DSTopPowFive{}, and \DSNeverFullPct{} of all cells never pass scientific acceptance across five trials; (3) task difficulty explains \DSVarTask{} of score variance, while swapping the harness on HY4 (\DSHarnessGap{} points, 95\% CI \DSHarnessCI{}) outweighs swapping models within a fixed harness; (4) inter-agent task correlations average \DSInterCorr{}, yet the Best-of-11 Oracle ceiling (\DSOracleMean{}, \DSOracleGain{} above \#1) reveals substantial complementarity alongside \DSUnsolvedAll{} universally unsolved frontier tasks; (5) longer runtime does not improve accuracy (Spearman $\rho=\DSRhoTime{}$), with timeouts concentrated in open-ended search such as \DSTopOutcomeTagEN{} (\DSTopOutcomeRate{}); and (6) capabilities diverge sharply across task types, with Qwen4-Max + Qoder (41.8) outperforming GPT-6.2 (34.5--36.4) on Tool-Use \& Planning.
\end{abstract}

\section{Introduction: Introducing DeltaScience}\label{sec:intro}

An AI agent hands back a scientific deliverable: plots are rendered and code runs to completion. The decisive question follows immediately---can this result actually be used? Does the conclusion survive a finer grid? Can a colleague reproduce the same numbers? In academic laboratories and industrial R\&D teams, only artifacts that pass rigorous scientific acceptance enter downstream decisions.

Most existing science benchmarks still ask ``how much does a model know?'' Answering static disciplinary questions does not establish whether an agent can select an appropriate solver, configure domain tools, execute long-horizon calculations, diagnose failures, and deliver verifiable artifacts. Recent workflow-oriented evaluations---such as Terminal-Bench~\cite{merrill2026terminalbench}, Terminal-Bench-Science~\cite{tbscience2026}, AstaBench~\cite{bragg2025astabench}, and Science Prediction Universe (SPU)~\cite{tu2026spu}---confirm a substantial gap between static recall and end-to-end scientific judgment.

We introduce \textbf{DeltaScience}\footnote{DeltaScience official platform and public leaderboard: \url{https://deltascience.bohrium.com}}, \textbf{an open evaluation platform for large models and AI agents in real scientific scenarios} built to bridge real-world academic research and industrial R\&D directly to AI agents. Guided by our core mission---\textbf{connecting frontier scientific discovery with real-world industrial R\&D, and testing AI scientific capability on problems that truly matter}---DeltaScience operationalizes a simple principle: \textbf{Real problems. Scientific rigor. Turning AI progress into scientific progress.}
\begin{itemize}
\item \textbf{Connecting frontier scientific discovery:} Partnering with active research groups to turn open scientific steps into tasks with explicit scientific goals and machine-verifiable acceptance rules.
\item \textbf{Connecting real-world industrial R\&D:} Partnering with drug discovery, materials, energy, and engineering teams to bring real physical constraints and operational trade-offs into evaluation, rather than substituting artificial difficulty for scientific value.
\end{itemize}

DeltaScience upholds five long-term commitments: \textbf{Real Problems} (scientific value with realistic constraints), \textbf{Scientific Rigor} (full-suite five-run evaluation on a pinned private benchmark with traceable protocols), \textbf{Understanding Boundaries} (dissecting capability limits across disciplines, task types, and failure modes without conflating computational scores with wet-lab success), \textbf{Visible Credit} (collective authorship, persistent CSTR identifiers, and citable records for contributors), and \textbf{Continuous Evolution} ($\delta$ embodies both verifiable scientific increments and continuous evolution: pinned-set retesting to measure true $\delta$ alongside periodic independent new-task expansion and staged public release of saturated tasks).

Because scientific outcomes depend jointly on reasoning and execution orchestration, DeltaScience evaluates versioned \textbf{Model + Harness} systems: the model supplies reasoning and generation, while the harness manages context, tool calls, code execution, and iterative refinement.

\begin{figure}[htbp]
\centering
\includegraphics[width=0.98\linewidth]{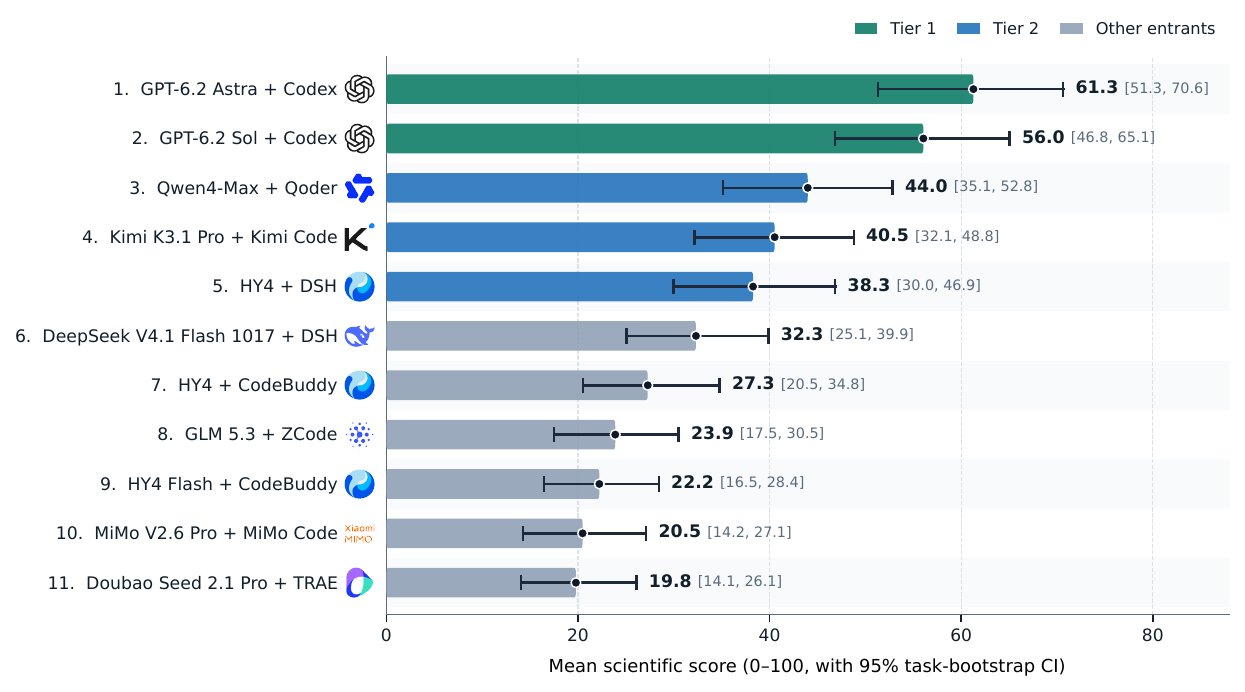}
\caption{\textbf{The inaugural \DSNumCombos{} Model + Harness configurations span a wide gradient across \DSNumTasks{} real scientific tasks, with even \#1 (\DSTopScore{}) leaving nearly 40 points of headroom.} Ordered descending by five-run mean with 95\% task-bootstrap CIs (five independent isolated runs per task; full statistics in Section~\ref{sec:results}).}
\label{fig:rank}
\end{figure}

Figure~\ref{fig:rank} highlights five headline findings: even the leading configuration (GPT-6.2 Astra + Codex, \DSTopScore{}) leaves nearly 40 points of headroom; \DSNeverFullPct{} of all task--configuration cells never pass scientific acceptance across five trials; swapping the harness on a fixed model (HY4: \DSHarnessGap{} points) outweighs swapping models within a fixed harness; the Best-of-11 Oracle ceiling (\DSOracleMean{}) reveals \DSOracleGain{} points of cross-system complementarity; and longer solving time correlates negatively with accuracy ($\rho=\DSRhoTime{}$).

The remainder of this report details benchmark task design and composition (Section~\ref{sec:benchmark}), evaluation methodology and 10,000-sandbox infrastructure (Section~\ref{sec:setup}), macro empirical results (Section~\ref{sec:results}), case studies and execution trace attribution (Section~\ref{sec:cases}), related work (Section~\ref{sec:related}), governance, future outlook, and open participation channels (Section~\ref{sec:community}), and concluding remarks (Section~\ref{sec:conclusion}).

\section{Benchmark Task Design and Composition}\label{sec:benchmark}

This section defines \emph{what} DeltaScience evaluates: the formal task specification illustrated via a public case study, the multi-label disciplinary taxonomy and cross-disciplinary structure of all \DSNumTasks{} v1.0 tasks, and the four mutually exclusive task types (evaluation methodology and sandbox infrastructure appear in Section~\ref{sec:setup}; governance and community roadmap appear in Section~\ref{sec:community}).

\subsection{Formal Task Specification and Public Case Study: FUS-0001}
Each task corresponds to a concrete, verifiable step extracted from active academic research or industrial R\&D. Before admission, every task must answer three questions: where the problem originates, what scientific objective it advances, and how its deliverable is verified. Formally, each task is defined by a four-tuple $\langle \mathcal{I}, \mathcal{C}, \mathcal{D}, \mathcal{V} \rangle$:
\begin{itemize}
\item \textbf{Inputs ($\mathcal{I}$):} Data, literature context, geometries, or initial boundary conditions provided to the agent.
\item \textbf{Execution constraints ($\mathcal{C}$):} Pinned domain software images, isolated container environment, compute allocation, time budget, and network policy.
\item \textbf{Required deliverables ($\mathcal{D}$):} Machine-checkable code, numerical fields, molecular/circuit/materials designs, or structured analytical reports.
\item \textbf{Scientific acceptance rules ($\mathcal{V}$):} An independent verifier computing domain-grounded scientific metrics mapped to $[0,1]$ continuous scores plus a pass threshold.
\end{itemize}
File existence or clean process exit never suffices; credit is earned strictly by scientific verification of the deliverable. The core benchmark remains private to prevent contamination, while three representative tasks---T14 (materials selective-release optimization), QI-0002 (noise-adapted quantum error-correcting circuits), and FUS-0001---are published on the website with scientific documentation and evaluation records.

\begin{figure}[htbp]
\centering
\includegraphics[width=0.72\linewidth]{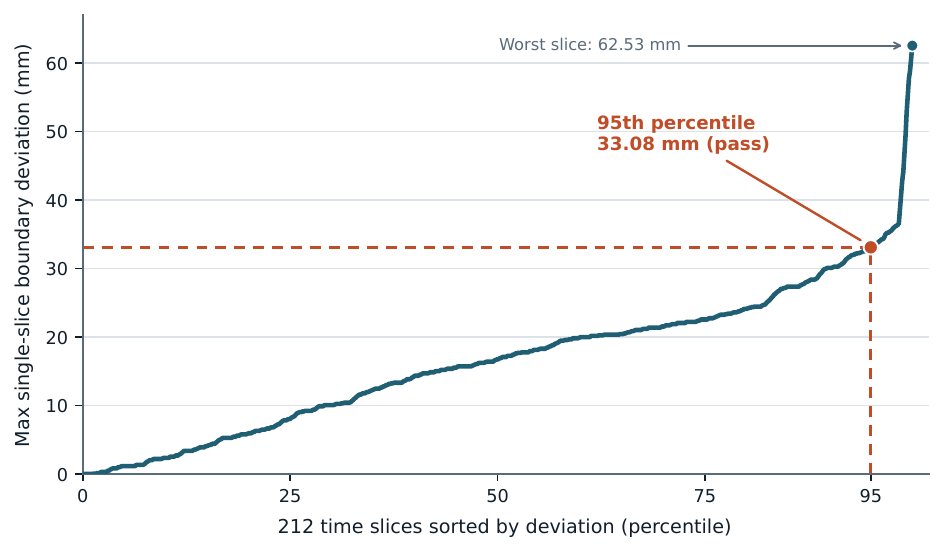}
\caption{\textbf{Passing submissions can still hide severe single-slice outliers, underscoring why full artifact archiving is essential for scientific review.} Per-slice plasma-boundary Hausdorff deviation across all 212 discharge slices for a passing HY4 submission on FUS-0001 (x-axis: 212 slices sorted ascending; y-axis: deviation in mm). The 95th percentile is 33.08\,mm (below the 33.83\,mm pass threshold, score 0.518; $\le 30$\,mm earns full credit), while the worst single slice still reaches 62.53\,mm.}
\label{fig:fus}
\end{figure}

As a concrete example, \textbf{FUS-0001 (Tokamak Free-Boundary Equilibrium)} tests plasma-boundary reconstruction from magnetic diagnostics. Following Pentland et al.'s validation of FreeGSNKE against EFIT++ on MAST-U discharge \#45425~\cite{pentland2025validation}, the task fixes machine geometry, coil and passive-conductor currents, and plasma profiles across 212 time slices, requiring the agent to deliver reproducible code and a $65\times65$ total poloidal flux field per slice. The independent verifier extracts the core plasma boundary, computes the maximum Hausdorff deviation over 360 angles per slice, and evaluates the 95th-percentile deviation $E$ (mm) against the archived FreeGSNKE reference $E_0=35.11$\,mm:
\[
S=\begin{cases}
\min\!\left(1,\;0.2+0.8\,\dfrac{E_0-E}{E_0-30}\right), & E\le E_0,\\[9pt]
\max\!\left(0,\;0.2\,\dfrac{50-E}{50-E_0}\right), & E>E_0.
\end{cases}
\]
Reproducing the baseline earns 0.20; $S\ge0.40$ ($E\le33.83$\,mm) passes; $E\le30$\,mm earns full credit. Reference boundaries live exclusively inside the verifier sandbox. Figure~\ref{fig:fus} shows a passing submission whose 95th-percentile error (33.08\,mm) clears the threshold while its worst single slice still reaches 62.53\,mm, illustrating why full artifact preservation is indispensable.

\FloatBarrier
\subsection{Multi-Label Disciplinary Taxonomy and Cross-Disciplinary Structure}\label{sec:composition}
DeltaScience v1.0 comprises \textbf{\DSNumTasks{} leaderboard tasks} (58 core tasks plus 7 BioMaster tasks under one unified protocol). Because real scientific problems are inherently interdisciplinary, each task receives \textbf{1--3 substantive disciplinary subfields}---permitting cross-group annotation while excluding generic numerical-tool inflation and catch-all ``other'' buckets (Table~\ref{tab:composition}).

\begin{table}[H]
\centering\small
\caption{\textbf{All \DSNumTasks{} v1.0 tasks span \DSNumGroups{} discipline groups and \DSNumSubfields{} subfields, with \DSCrossCount{} tasks (\DSCrossPct{}) crossing two or more groups.} Each task receives 1--3 substantive disciplinary subfields without catch-all buckets.}
\label{tab:composition}
\begin{tabularx}{\linewidth}{@{}l r X@{}}
\toprule
Discipline Group & Tasks & Subfields Covered (35 total) \\
\midrule
Chemistry & 24 & Medicinal Chem. (17), Organic Chem. (13), Cheminformatics (5), Comp. Chem. (4), Chemical Bio. (2), Mol. Spectroscopy (2), Catalysis \& Electrochem. (1), Analytical Chem. (1) \\
Physics & 22 & Quantum Comp. (8), Fusion \& Plasma (5), Condensed Matter (5), Stat. Physics (2), AMO Physics (1), Biophysics (1) \\
Life Sciences & 21 & Omics (13), Structural Bio. (7), Bioinformatics (5), Immunology \& Inflamm. (3), Cancer Biology (3), Stem Cell \& Aging (2) \\
Math \& CS & 11 & Algorithm Design (4), Algebra \& Comb. (3), Operations Res. (3), Machine Learning (3), Control Theory (1), Coding Theory (1), Comp. Math (1) \\
Materials Science & 9 & Comp. Materials (5), Surface \& Nanomat. (3), Materials Design (1), Mech. of Materials (1) \\
Engineering & 6 & Eng. Design (4), Fluid Mech. (2), Heat Transfer (1) \\
Earth Science & 1 & Geophysics (1) \\
\bottomrule
\end{tabularx}

\end{table}

\begin{figure}[H]
\centering
\includegraphics[width=0.70\linewidth]{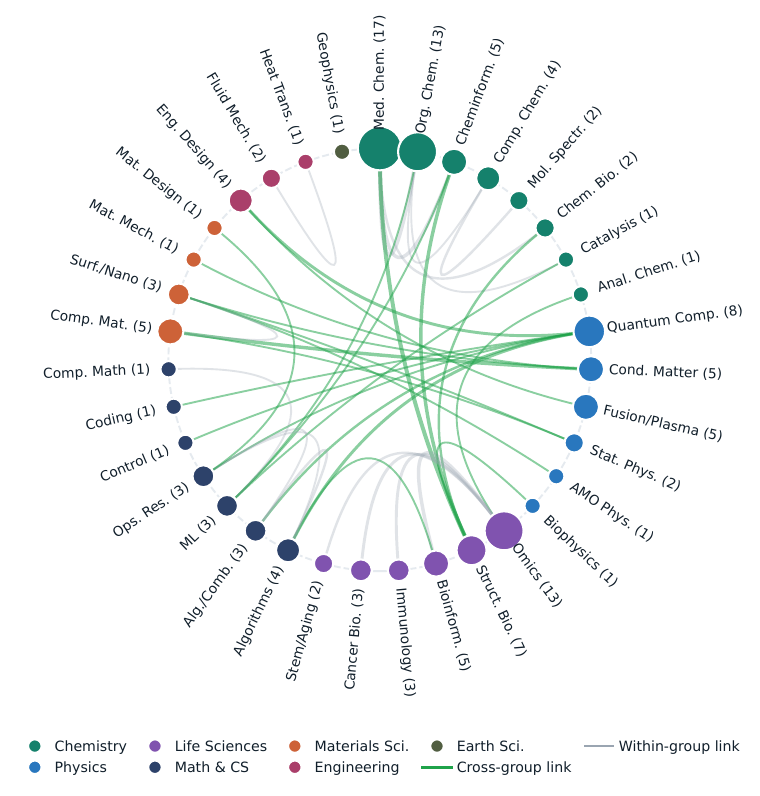}
\caption{\textbf{The research-direction network reveals dense interdisciplinary hubs connecting Chemistry, Life Sciences, Physics, and Materials Science.} Co-occurrence graph of all 35 subfields across 7 discipline groups (node size scales with task count; green arcs denote cross-group links and gray arcs within-group co-occurrences).}
\label{fig:network}
\end{figure}

Figure~\ref{fig:network} maps the subfield co-occurrence structure across all \DSNumSubfields{} research directions: Chemistry, Life Sciences, Physics, and Materials Science form dense cross-group hubs, mirroring the interdisciplinary coupling of active scientific research.

\FloatBarrier
\subsection{Mutually Exclusive Task Types}\label{sec:tasktypes}
Independently, each task belongs to exactly one of \textbf{4 mutually exclusive task types} based on its deliverable and verification structure (Table~\ref{tab:tasktypes}), separating agent capabilities by execution paradigm. Appendix~\ref{app:tasklist} provides the sanitized directory of all \DSNumTasks{} tasks.

\begin{table}[htbp]
\centering\small
\caption{\textbf{Four mutually exclusive task types partition all \DSNumTasks{} tasks across design, analysis, simulation, and tool-use planning.} Task counts, deliverable structures, and representative scientific domains in DeltaScience v1.0.}
\label{tab:tasktypes}
\begin{tabularx}{\linewidth}{@{}l r X@{}}
\toprule
Task Type & Tasks & Defining Characteristics and Examples \\
\midrule
Open Design \& Search & 23 & No unique answer; agent submits a molecular, circuit, or coil design graded continuously on physical performance (e.g., molecular glues, stellarator boundaries, quantum codes). \\
Multi-Stage Analysis & 16 & Multi-cohort or multi-omics pipelines requiring evidence-backed mechanistic conclusions (e.g., single-cell and tumor heterogeneity analyses). \\
Simulation \& Comp. & 15 & Solver-intensive numerical calculations checked against reference fields or invariants (e.g., tokamak equilibrium, supersonic boundary layers, 2D material excitations). \\
Tool-Use \& Planning & 11 & Iterative invocation of specialized domain tools to synthesize and validate complete routes (e.g., open-route retrosynthesis). \\
\bottomrule
\end{tabularx}
\end{table}
\FloatBarrier

\section{Evaluation Methodology and Large-Scale Sandbox Infrastructure}\label{sec:setup}

Credible scientific agent evaluation requires both rigorous statistical protocols and an isolated, auditable execution substrate capable of sustaining thousands of long-horizon runs. This section details \emph{how} DeltaScience evaluates: five-run scoring and statistical inference, controlled Model + Harness hosting, immutable versioning with three-layer nested sandboxes, and 10,000+ sandbox concurrent scheduling via EvalWheel.

\subsection{Scoring Protocol, Multi-Run Reliability, and Statistical Inference}\label{sec:protocol}
\textbf{Five-run equal-weight aggregation.} Let $s_{iar}\in[0,1]$ denote the continuous score of configuration $a$ on task $i\in\{1,\dots,N\}$ ($N=\DSNumTasks{}$) in independent trial $r\in\{1,\dots,5\}$. We average the five trials within each task before taking the unweighted mean across tasks on a 0--100 scale:
\[
m_{ia}=\frac{1}{5}\sum_{r=1}^{5}s_{iar},\qquad
S_a=\frac{100}{N}\sum_{i=1}^{N}m_{ia}.
\]
This estimator captures expected single-attempt scientific delivery rather than best-of-five luck.

\textbf{Multi-run reliability (pass@1, pass@5, pass\textsuperscript{5}).} Alongside the continuous score $s_{iar}\in[0,1]$, each task records a binary scientific acceptance outcome $p_{iar}\in\{0,1\}$ indicating whether the deliverable clears that task's scientific pass threshold (e.g., $S\ge 0.40$ on FUS-0001). To distinguish occasional success from engineering-grade dependability, let $c_{ia}=\sum_{r=1}^{5}p_{iar}$ count passing trials across five runs. Following Chen et al.~\cite{chen2021codex} and $\tau$-bench~\cite{yao2024taubench}, we report:
\[
\text{pass@1}_a=\frac{1}{N}\sum_{i=1}^{N}\frac{c_{ia}}{5},\quad
\text{pass@5}_a=\frac{1}{N}\sum_{i=1}^{N}\mathbb{I}[c_{ia}\ge 1],\quad
\text{pass}^5_a=\frac{1}{N}\sum_{i=1}^{N}\mathbb{I}[c_{ia}=5].
\]

\textbf{Task-level paired bootstrap and variance decomposition.} Because intrinsic task difficulty spans the full $[0,100]$ range, independent-sample tests severely understate true discriminative power. For every single-system mean $S_a$ and pairwise contrast $\Delta_{ab}=S_a-S_b=\frac{100}{N}\sum_{i=1}^{N}(m_{ia}-m_{ib})$, we draw \textbf{5,000 task-level bootstrap resamples with replacement} and take the 2.5th and 97.5th percentiles as 95\% confidence intervals. Crucially, these two intervals measure distinct uncertainties: a single system's 95\% CI captures \emph{task-composition sensitivity} (how much its mean would vary on another 65-task draw from the same distribution, hence $\approx\pm 9$--$10$ points wide---far broader than five-run execution noise on the fixed suite), whereas pairwise tests use \emph{task-paired} differences that cancel shared task difficulty. Consequently, two systems with overlapping single-system CIs can still be cleanly separated whenever the paired 95\% CI of $\Delta_{ab}$ excludes zero. We additionally perform two-way main-effect variance decomposition ($\text{SS}_{\text{total}}=\text{SS}_{\text{task}}+\text{SS}_{\text{combo}}+\text{SS}_{\text{resid}}$) over all cell means $m_{ia}$ (full formulas in Appendix~\ref{app:stats}).

\textbf{Time-bounded execution and scoring of timeouts, refusals, and harness errors.} Every task enforces a unified, research-grounded per-task time budget calibrated to its computational scope: across all \DSNumTasks{} tasks, limits span \textbf{\DSTimeLimitMin{}--\DSTimeLimitMax{} hours} with a \textbf{median of \DSTimeLimitMed{} hours} (mean \DSTimeLimitMean{} hours), and \textbf{\DSTimeLimitGeEightN{} tasks (\DSTimeLimitGeEightPct{})} grant at least 8 hours per run (full limits in Appendix~\ref{app:tasklist}, Table~\ref{tab:tasklist}). Budget-exhausted timeouts, explicit model refusals, and harness errors strictly score 0 and remain included in all five-trial averages and reliability metrics. Strict time limits rest on two core principles:
\begin{itemize}
\item \textbf{Real-world research usability:} Latency and compute cost are integral to scientific and industrial R\&D, where researchers cannot wait indefinitely or absorb runaway token bills for one step.
\item \textbf{Curbing overthinking and brute-force loops:} Removing limits rewards unchecked retries and masks weaknesses in budget-aware convergence. Empirically, under the \DSTimeLimitMed{}-hour median budget, Tier-1 systems (GPT-6.2 Sol / Astra + Codex) incur only 3\%--5\% timeouts while solving tasks in a median 35--42 minutes ($<1/10$ of the median limit) using 105k--126k output tokens, confirming the budgets are generous rather than artificially restrictive.
\end{itemize}

\begin{figure}[htbp]
\centering
\begin{tikzpicture}[x=18mm,y=0.36mm]
\foreach \y in {25,50,75,100}{
  \draw[PKRule,densely dashed,line width=0.45pt](0.35,\y)--(5.65,\y);
  \node[pknote,anchor=east]at(0.28,\y){\y};}
\draw[PKMuted!65,line width=0.65pt](0.35,0)--(5.65,0);
\node[pknote,anchor=east]at(0.28,0){0};
\draw[PKAmber,densely dotted,line width=0.85pt](0.35,40)--(5.65,40);
\node[font=\scriptsize,text=PKAmber,anchor=west]at(5.72,40){Pass line 40.0};
\draw[PKTeal,densely dashed,line width=0.85pt](0.35,72.8)--(5.65,72.8);
\node[font=\scriptsize\bfseries,text=PKTeal,anchor=west]at(5.72,72.8){5-run mean 72.8};
\foreach \x/\v/\col in {1/100/PKAccent!82,2/100/PKAccent!82,3/64.2/PKBlue!75,5/100/PKAccent!82}{
  \fill[\col,rounded corners=1.5pt](\x-0.24,0)rectangle(\x+0.24,\v);
  \node[fill=white,inner xsep=2pt,inner ysep=0.5pt,font=\footnotesize\bfseries,text=PKInk,anchor=south]at(\x,\v+2){\v};}
\fill[PKCoral](4-0.24,0)rectangle(4+0.24,3.5);
\node[fill=PKCoral!10,draw=PKCoral!60,rounded corners=2pt,inner xsep=4pt,inner ysep=1.5pt,
  font=\scriptsize\bfseries,text=PKCoral,anchor=south]at(4,10){0 (timeout)};
\foreach \x in {1,2,3,4,5}{\node[pknote,anchor=north]at(\x,-5){Run \x};}
\node[pknote,anchor=south west]at(0.35,109){Score (0--100)};
\end{tikzpicture}
\caption{\textbf{A single timeout pulls a full-scoring system's five-run mean down to 72.8, illustrating why single-trial benchmarks mask operational volatility.} Five-run score trajectory of HY4 + CodeBuddy on FUS-0001: three full scores, one partial score (64.2), and one timeout (0).}
\label{fig:fiveruns}
\end{figure}
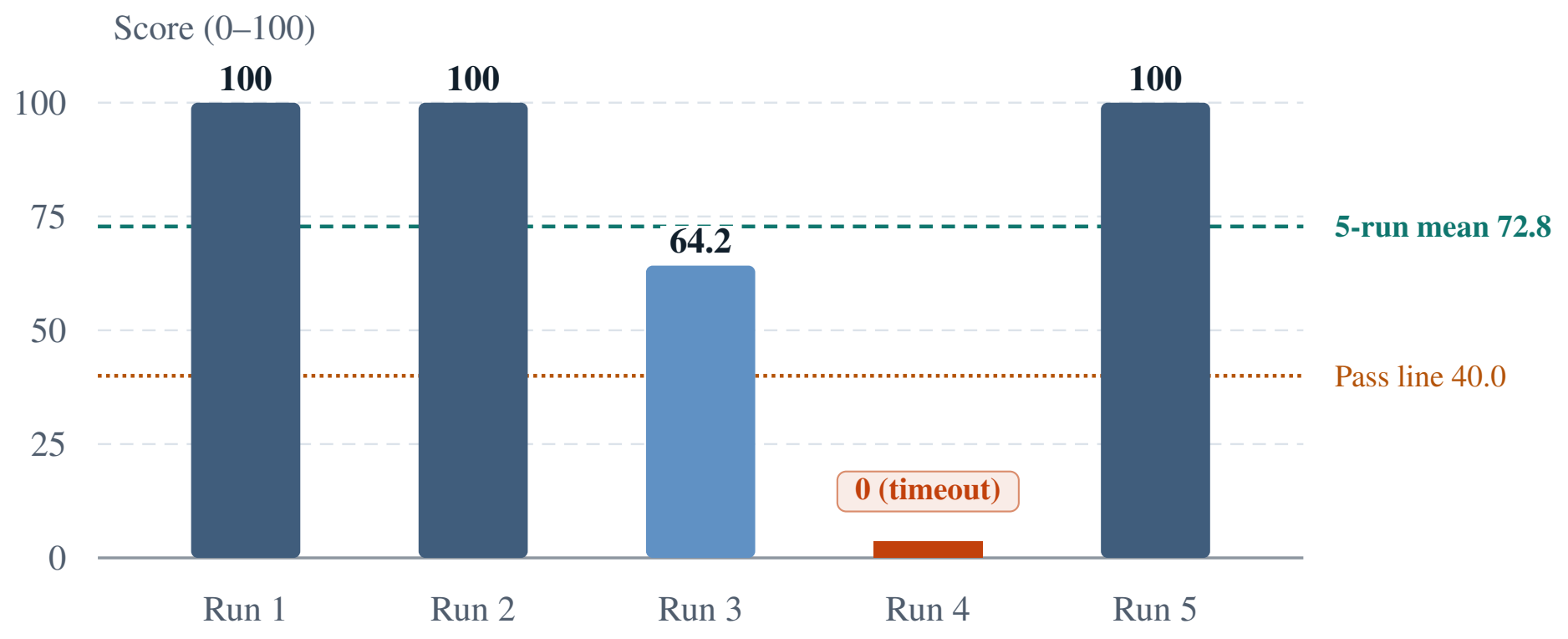

As Figure~\ref{fig:fiveruns} illustrates, the same system on the same task can span three full scores, one partial score (64.2), and one timeout (0), demonstrating why single-trial evaluation fails to capture dependable scientific delivery.

\FloatBarrier
\subsection{Controlled Configuration Matrix and Model Hosting}
The inaugural v1.0 release evaluates \textbf{\DSNumCombos{} Model + Harness configurations} spanning 10 model versions and 8 harnesses (Table~\ref{tab:configs}). Four built-in controlled contrasts---HY4 across DSH and CodeBuddy, two models inside DSH, two models inside CodeBuddy, and GPT-6.2 Astra vs.\ Sol inside Codex---enable direct separation of model and harness effects on identical tasks.

\begin{table}[htbp]
\centering\small
\caption{\textbf{The inaugural \DSNumCombos{} configurations span 10 models and 8 harnesses with four built-in controlled Model vs.\ Harness contrasts.} Complete roster of evaluated Model + Harness pairs in DeltaScience v1.0.}
\label{tab:configs}
\begin{tabular}{@{}ll@{\qquad\qquad}ll@{}}
\toprule
Model & Harness & Model & Harness \\
\midrule
\modellogo{gpt}GPT-6.2 Astra & Codex & \modellogo{hy4}HY4 & DSH \\
\modellogo{gpt}GPT-6.2 Sol & Codex & \modellogo{deepseek}DeepSeek V4.1 Flash 1017 & DSH \\
\modellogo{qwen}Qwen4-Max & Qoder & \modellogo{hy4}HY4 & CodeBuddy \\
\modellogo{glm}GLM 5.3 & ZCode & \modellogo{hy4}HY4 Flash & CodeBuddy \\
\modellogo{doubao}Doubao Seed 2.1 Pro & TRAE & \modellogo{kimi}Kimi K3.1 Pro & Kimi Code \\
\modellogo{mimo}MiMo V2.6 Pro & MiMo Code & & \\
\bottomrule
\end{tabular}
\end{table}

\textbf{Self-hosted GPU platform, weight pinning, and gateway auditing.} Underlying GPU compute and inference serving are powered by \textbf{Trisol}: it hosts weight-, stack-, and parameter-pinned self-hosted models for agent inference calls while also supplying and scheduling the underlying GPU compute consumed by scientific sandboxes during solving and verification. Other evaluated models connect via official first-party APIs through a unified model gateway logging per-request time-to-first-token, end-to-end latency, and input/output/cached token counts for exact reconciliation. Because long-context agent concurrency is bounded by GPU KV-cache throughput per replica (e.g., $\sim$48 concurrent slots for Kimi K3.1 Pro, $\sim$64 for DeepSeek V4.1 Flash 1017), scheduling enforces \textbf{dual-bottleneck admission} across both sandbox/GPU cluster quotas and model replica capacity.

\subsection{Immutable Versioning and Three-Layer Isolated Sandbox Architecture}
Evaluations execute on Bohrium's modular infrastructure (Table~\ref{tab:platform}) under two invariants: \textbf{immutable versioning} and \textbf{physical agent--verifier isolation}. Every task is registered as a Flywheel Asset, assembled into a pinned Bundle (\texttt{ID@vN}), and persisted in the Wenyon immutable object store. Every evaluation record binds exact task snapshots, container image digests, and archived deliverables via content hashes for drift-free retesting.

\begin{table}[htbp]
\centering\small
\caption{\textbf{Seven modular infrastructure components decouple version locking, isolated execution, and 10,000+ sandbox scheduling.} Core components and responsibilities of the DeltaScience evaluation stack.}
\label{tab:platform}
\begin{tabularx}{\linewidth}{@{}p{26mm}X@{}}
\toprule
Component & Responsibility \\
\midrule
Wenyon & Immutable object store for task packages, deliverables, full execution traces, and access control. \\
Flywheel & Versioned asset and bundle registry pinning immutable snapshots (\texttt{ID@vN}). \\
Bohrium Sandbox & Isolated cloud container environments for orchestration, solving, and grading. \\
Harbor LBG & Per-task runner launching harnesses, collecting deliverables, and invoking verifiers. \\
EvalWheel & Batch scheduler governing whole-task admission, dual-bottleneck queuing, archiving, and cleanup. \\
Model Gateway & Unified inference routing with per-session latency and token accounting. \\
Trisol & Underlying GPU compute and serving platform powering both agent model inference and sandbox GPU scheduling. \\
\bottomrule
\end{tabularx}
\end{table}

\begin{figure}[htbp]
\centering
\includegraphics[width=0.98\linewidth]{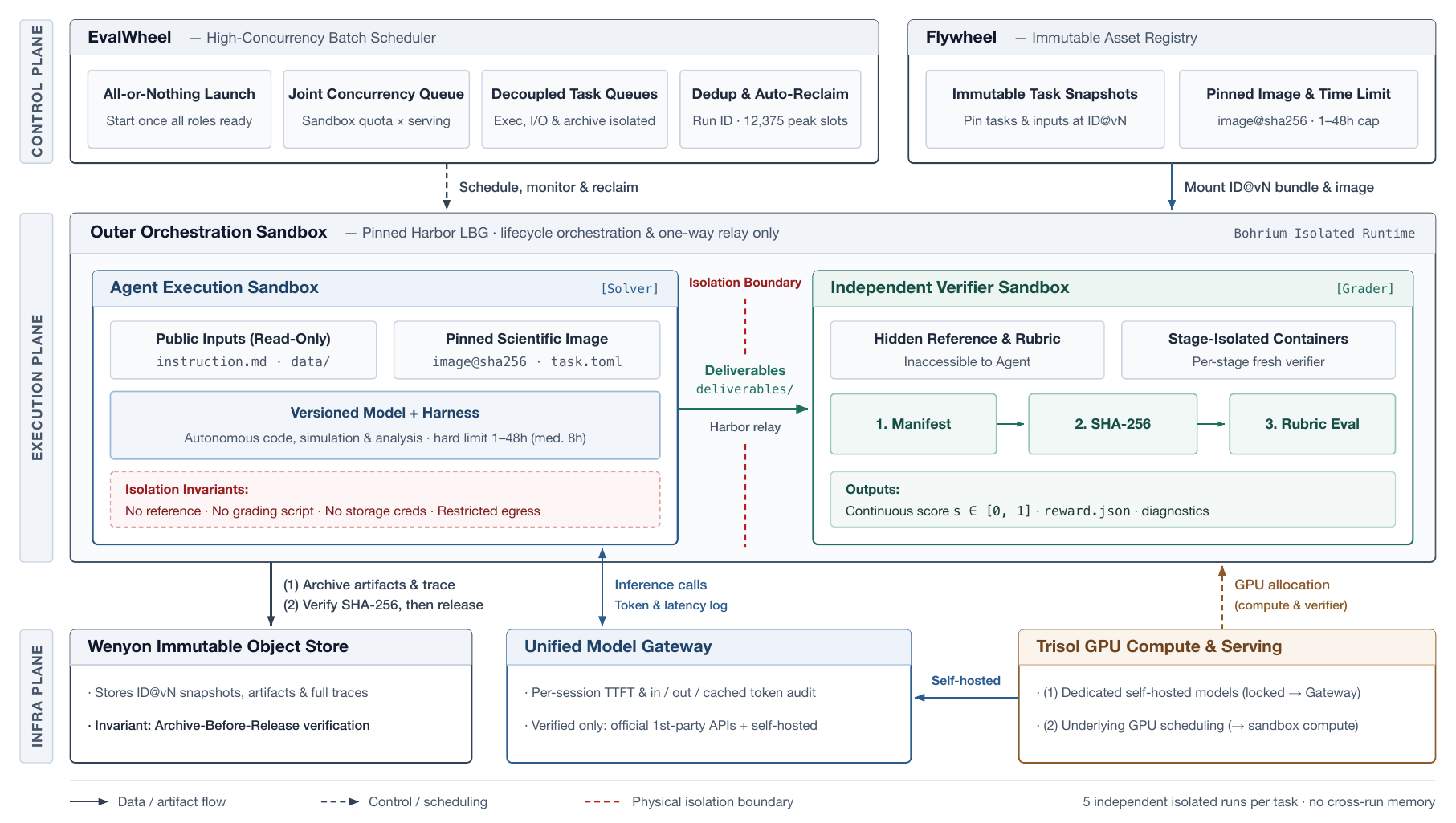}
\caption{\textbf{DeltaScience three-layer isolated sandbox and infrastructure architecture.} EvalWheel orchestrates 10,000+ concurrent slots while Flywheel locks immutable \texttt{ID@vN} task snapshots and container digests; inside Bohrium, pinned Harbor LBG physically isolates the Agent sandbox from the independent Verifier sandbox; underneath, Wenyon enforces archive-and-verify-before-release, the Unified Model Gateway meters inference and token usage across official first-party APIs and self-hosted models, and Trisol provides both locked self-hosted model serving and underlying GPU compute for sandboxes.}
\label{fig:sandbox}
\end{figure}

Each trial launches a \textbf{three-layer parent--child sandbox} (Figure~\ref{fig:sandbox}):
\begin{enumerate}
\item \textbf{Outer orchestration sandbox:} Runs a pinned Harbor LBG instance strictly for lifecycle orchestration and isolated artifact relay;
\item \textbf{Agent sandbox:} Mounts the task's pinned domain software image and public inputs $\mathcal{I}$, physically stripping reference answers, grading scripts, and storage credentials under task-locked time and network policies;
\item \textbf{Independent Verifier sandbox:} Receives deliverables one-way from the outer container (using separate containers per stage on multi-stage tasks), verifies the delivery manifest and recomputes SHA-256 content hashes before computing continuous scientific scores.
\end{enumerate}
Teardown enforces a strict \textbf{archive-before-release invariant}: all deliverables, interaction logs, and execution traces are persisted to Wenyon and verified by hash pull-back before any container is reclaimed.

\FloatBarrier
\subsection{Large-Scale Concurrent Scheduling: 10,000+ Sandbox Evaluation Substrate}
As an open evaluation platform for large models and AI agents serving full-suite retesting, year-round Playground authoring competitions, and periodic new-task releases, DeltaScience sustains massive concurrent throughput without sacrificing isolation or auditability. Because each trial reserves an outer orchestration sandbox plus agent and independent verifier sandboxes (isolated per stage on multi-stage tasks), peak batch reservation reaches \textbf{12,375 concurrent sandbox slots}---exceeding ten thousand sandboxes simultaneously.

Orchestrating \textbf{Wenyon + Flywheel + Bohrium three-layer isolated sandboxes + EvalWheel}, the platform guarantees deterministic execution at 10,000+ sandbox scale via five systems mechanisms:
\begin{itemize}
\item \textbf{Whole-task dual-bottleneck admission:} Admits a trial only when all sandbox roles and corresponding Trisol model replica slots are available, preventing partial-allocation deadlocks via smooth backoff queuing.
\item \textbf{Adaptive creation concurrency window:} Shares a dynamic creation window (initialized at 16 concurrent launches) per project and spec, throttling only container creation while leaving running trials unaffected.
\item \textbf{Idempotent instance lifecycle:} Reconciles container state against unique request IDs so every trial maps to exactly one isolated instance without duplicates or orphan containers.
\item \textbf{Six-lane operation isolation and image prewarming:} Separates control, execution, upload, download, archive, and packaging into non-blocking queues while binding all roles to prewarmed container templates.
\item \textbf{Full trace archiving and hash verification:} Persists all interaction logs, tool traces, and deliverables to Wenyon with pull-back digest checks prior to container reclamation, guaranteeing post-hoc auditability for every score and failure mode.
\end{itemize}

\section{Macro Empirical Results and Quantitative Analysis}\label{sec:results}

This section analyzes all \textbf{\DSNumRuns{} independent sandbox runs} across the \DSNumTasks{} DeltaScience v1.0 tasks and \DSNumCombos{} Model + Harness configurations along five macro dimensions: overall ranking and multi-run reliability, controlled Model vs.\ Harness decomposition, inter-agent agreement and Oracle complementarity, time/token Pareto efficiency and failure modes, and multi-dimensional task-type and discipline profiles (execution trace attribution and microscopic case studies follow in Section~\ref{sec:cases}).

\subsection{Overall Ranking, Multi-Run Reliability, and Pairwise Dominance}
Table~\ref{tab:rank} summarizes the inaugural leaderboard across all \DSNumCombos{} configurations, pairing five-run mean scores and 95\% bootstrap CIs with multi-run reliability (\textbf{pass@1}, \textbf{pass@5}, \textbf{pass\textsuperscript{5}}), timeout/error rates, and compute efficiency.

\begin{table}[htbp]
\centering\scriptsize
\caption{\textbf{GPT-6.2 systems lead Tier 1, yet even the top configuration leaves nearly 40 points of headroom on real scientific deliverables.} DeltaScience v1.0 (\DSNumTasks{} tasks) overall leaderboard, multi-run reliability, and resource usage (0--100, ordered descending by five-run continuous mean). Intervals are 5,000-draw task-bootstrap 95\% CIs; under per-task binary (0/1) scientific acceptance criteria, pass@1 is expected single-run pass rate, pass@5 is the share of tasks passed in $\ge 1$ of 5 runs (capability ceiling), and pass\textsuperscript{5} is the share passed in all 5 runs (reliability floor); Time is median solving minutes; Eff.\ is score per million total tokens.}
\label{tab:rank}
\begin{tabular}{@{}r l r c r r r r r r r@{}}
\toprule
\# & Model + Harness & Mean & 95\% CI & pass@1 & pass@5 & pass\textsuperscript{5} & Timeout & Time & OutTok & Eff. \\
\midrule
1 & \modellogo{gpt}GPT-6.2 Astra + Codex & \textbf{61.3} & [51.3, 70.6] & 55.7\% & 60.0\% & 49.2\% & 4.9\% & 42m & 105k & 4.97 \\
2 & \modellogo{gpt}GPT-6.2 Sol + Codex & \textbf{56.0} & [46.8, 65.1] & 46.2\% & 53.8\% & 36.9\% & 3.1\% & 35m & 126k & 3.68 \\
3 & \modellogo{qwen}Qwen4-Max + Qoder & 44.0 & [35.1, 52.8] & 33.2\% & 49.2\% & 20.0\% & 12.4\% & 176m & 201k & 1.25 \\
4 & \modellogo{kimi}Kimi K3.1 Pro + Kimi Code & 40.5 & [32.1, 48.8] & 29.5\% & 50.8\% & 15.4\% & 8.6\% & 150m & 217k & 1.11 \\
5 & \modellogo{hy4}HY4 + DSH & 38.3 & [30.0, 46.9] & 24.3\% & 44.6\% & 10.8\% & 6.2\% & 148m & 216k & 1.29 \\
6 & \modellogo{deepseek}DeepSeek V4.1 Flash 1017 + DSH & 32.3 & [25.1, 39.9] & 17.8\% & 40.0\% & 4.6\% & 8.2\% & 191m & 228k & 0.88 \\
7 & \modellogo{hy4}HY4 + CodeBuddy & 27.3 & [20.5, 34.8] & 15.1\% & 29.2\% & 4.6\% & 14.9\% & 172m & 291k & 1.17 \\
8 & \modellogo{glm}GLM 5.3 + ZCode & 23.9 & [17.5, 30.5] & 15.4\% & 36.9\% & 3.1\% & 32.3\% & 300m & 211k & 0.90 \\
9 & \modellogo{hy4}HY4 Flash + CodeBuddy & 22.2 & [16.5, 28.4] & 9.8\% & 21.5\% & 1.5\% & 5.2\% & 114m & 107k & 1.17 \\
10 & \modellogo{mimo}MiMo V2.6 Pro + MiMo Code & 20.5 & [14.2, 27.1] & 9.2\% & 16.9\% & 4.6\% & 13.6\% & 159m & 167k & 1.03 \\
11 & \modellogo{doubao}Doubao Seed 2.1 Pro + TRAE & 19.8 & [14.1, 26.1] & 10.8\% & 24.6\% & 1.5\% & 37.5\% & 300m & 160k & 0.51 \\
\bottomrule
\end{tabular}

\end{table}

Clear performance tiers emerge across the field: Codex-backed \textbf{GPT-6.2 Astra (\DSTopScore{})} and \textbf{GPT-6.2 Sol (\DSSecondScore{})} anchor Tier 1; \textbf{Qwen4-Max + Qoder (\DSThirdScore{})}, \textbf{Kimi K3.1 Pro + Kimi Code (\DSFourthScore{})}, and \textbf{HY4 + DSH (\DSFifthScore{})} form a competitive Tier 2; and the remaining six configurations span 19.8--32.3.

\begin{figure}[htbp]
\centering
\includegraphics[width=0.92\linewidth]{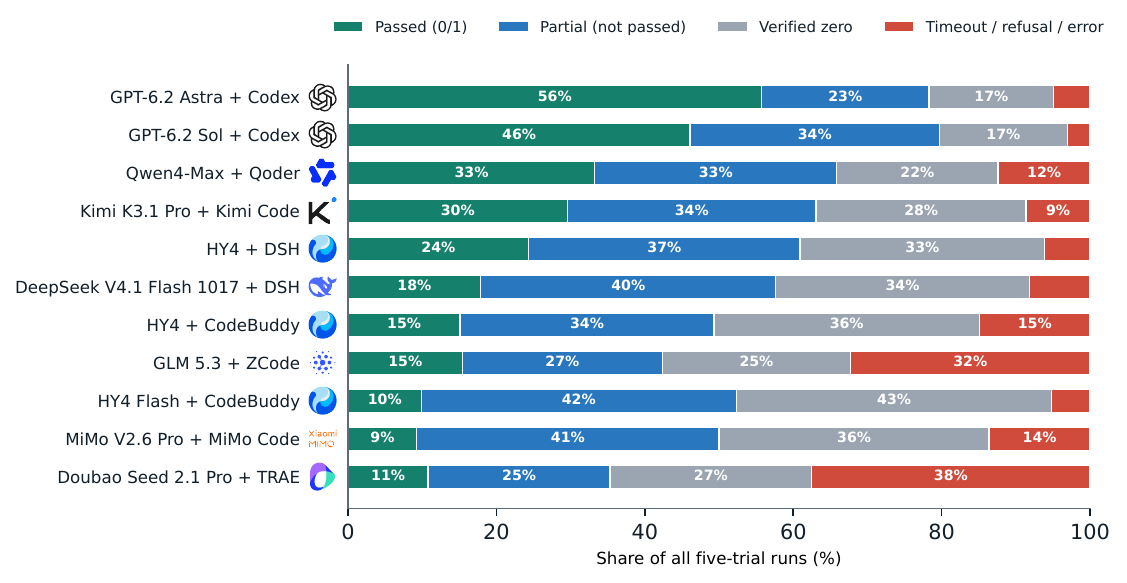}
\caption{\textbf{Tier-1 systems pass scientific acceptance on 46\%--56\% of runs, whereas over one-third of runs in high-timeout systems terminate in timeouts or errors.} Four-way run outcome composition across all five trials per configuration (passed 0/1, partial not passed, verified zero, and timeout/refusal/error).}
\label{fig:outcomes_stack}
\end{figure}

Decomposing individual trial outcomes (Figure~\ref{fig:outcomes_stack}) shows that rank separation reflects both scientific pass rates and execution convergence. Tier-1 systems pass acceptance on 46\%--56\% of trials and Tier 2 spans 24\%--33\%, whereas 32\%--38\% of trials for GLM 5.3 + ZCode and Doubao Seed 2.1 Pro + TRAE end in timeouts or errors, directly depressing their overall scores.

\begin{figure}[htbp]
\centering
\begin{tikzpicture}[x=18mm,y=0.52mm]
\foreach \y in {20,40,60,80}{\draw[PKRule,densely dashed,line width=0.45pt](-0.1,\y)--(6.1,\y);\node[pknote,anchor=east]at(-0.18,\y){\y\%};}
\draw[PKMuted!65,line width=0.65pt](-0.1,0)--(6.1,0);
\node[pknote,anchor=east]at(-0.18,0){0\%};
\node[fill=PKPaper,draw=PKRule,rounded corners=2pt,inner xsep=5pt,inner ysep=2.5pt,font=\scriptsize,text=PKMuted,anchor=north east]at(6.05,78){Median within-task SD 4.0 pts};
\foreach \i/\pct/\col in {
  0/61.1/PKCoral!80,
  1/9.8/PKSlate!52,
  2/6.7/PKSlate!52,
  3/5.2/PKSlate!52,
  4/3.4/PKSlate!52,
  5/13.8/PKTeal!82}{
  \fill[\col,rounded corners=1.5pt](\i+0.18,0)rectangle(\i+0.82,\pct);
  \node[font=\scriptsize\bfseries,text=PKInk,anchor=south]at(\i+0.5,\pct+1.2){\pct\%};
  \node[pknote,anchor=north]at(\i+0.5,-2.5){\i/5};}
\node[pknote,anchor=north]at(3,-12.5){Passed trials across 5 runs (share of task--configuration cells)};
\end{tikzpicture}
\caption{\textbf{Across all task--configuration cells, \DSNeverFullPct{} never pass scientific acceptance across five trials and only \DSAllFiveFullPct{} achieve 5/5 reliability.} Distribution of passed trial counts (0--5 out of 5 runs under per-task 0/1 acceptance criteria) over all $65\times 11$ cells.}
\label{fig:stability}
\end{figure}
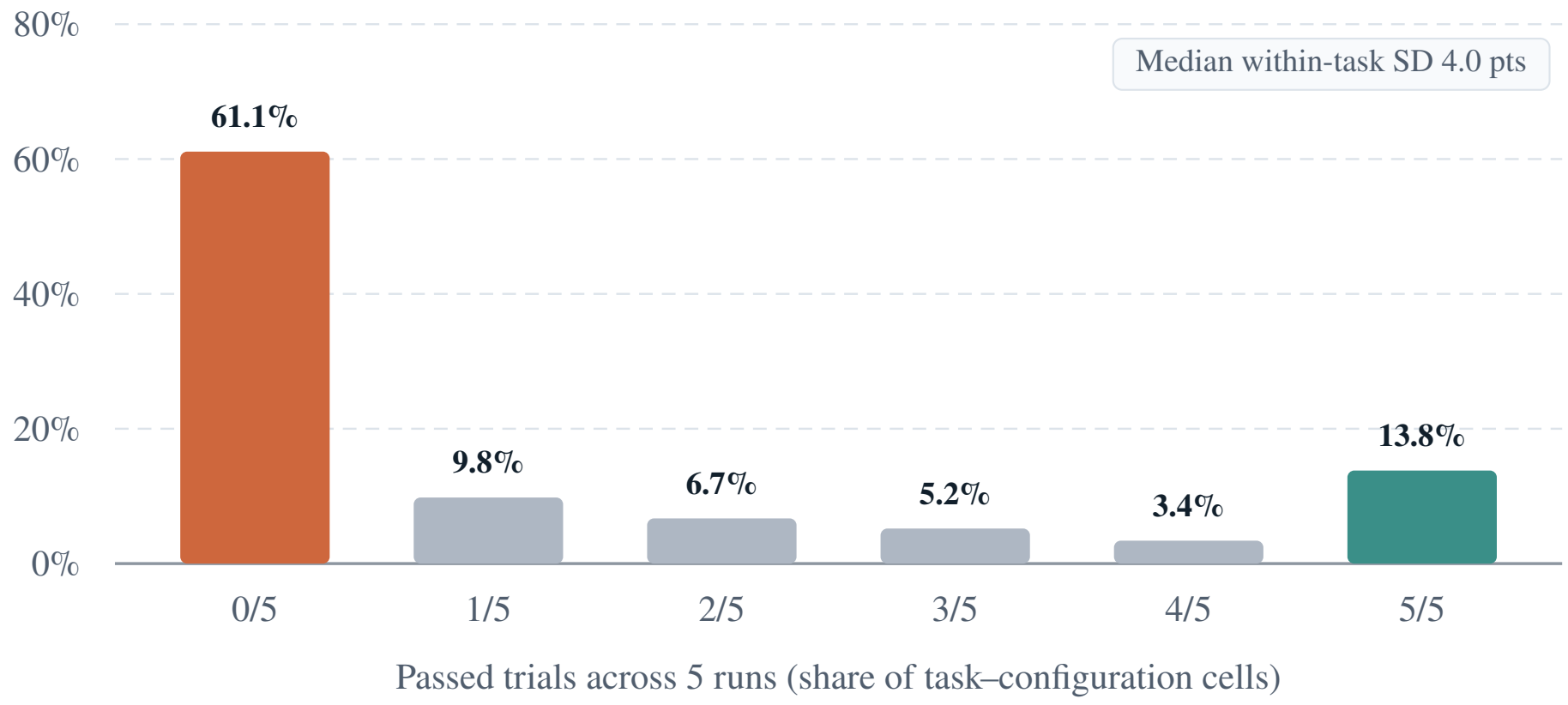

\textbf{A wide gap separates occasional success from dependable delivery (Figure~\ref{fig:stability}).} Even for \#1 GPT-6.2 Astra + Codex, pass@5 reaches \DSTopPassFive{} while strict 5-of-5 pass\textsuperscript{5} drops to \DSTopPowFive{}---a $>10$-point reliability penalty; inside Tier 2, Kimi K3.1 Pro + Kimi Code reaches 50.8\% pass@5 (slightly ahead of Qwen4-Max + Qoder at 49.2\%) yet falls to 15.4\% pass\textsuperscript{5} (vs.\ 20.0\% for Qwen4-Max). Although the low median within-cell SD (\DSMedWithinSD{} points) confirms stable ranking, single-trial benchmarks systematically overstate real-world scientific reliability.

\begin{figure}[H]
\centering
\includegraphics[width=0.80\linewidth]{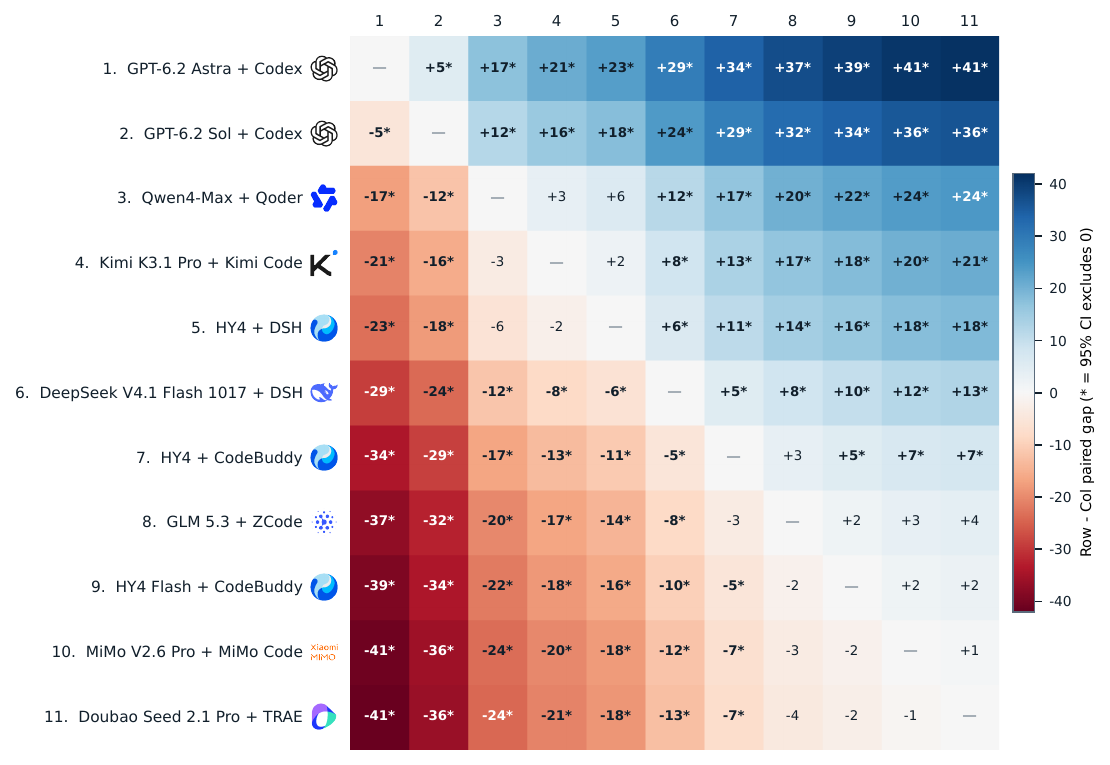}
\caption{\textbf{\DSPairSep{} of \DSPairTotal{} configuration pairs (81.8\%) are statistically separated, confirming strong discriminative power.} 11 $\times$ 11 pairwise score gap matrix (row minus column; * denotes 5,000-draw task-bootstrap 95\% CI excluding zero).}
\label{fig:pairwise}
\end{figure}

The pairwise dominance matrix (Figure~\ref{fig:pairwise}) illustrates why pairwise significance must be judged by task-paired tests rather than visual overlap of Table~\ref{tab:rank}'s single-system task-composition CIs: once shared task difficulty is cancelled out per task, cross-tier and most within-tier pairs achieve statistically significant separation (aside from a few adjacent Tier-2 pairs such as Qwen4-Max vs.\ Kimi K3.1 Pro and Kimi K3.1 Pro vs.\ HY4 + DSH).

\FloatBarrier
\subsection{Variance Decomposition and Controlled Model vs.\ Harness Contrasts}
Two-way main-effect variance decomposition across all cell means attributes \textbf{\DSVarTask{} of total variance to task difficulty} and \textbf{\DSVarCombo{} to configuration}, leaving \textbf{\DSVarResid{}} in task--configuration interaction and residual variance. While wide task-difficulty spread anchors ranking sensitivity, the large interaction component signals genuine specialization across scientific workflows.

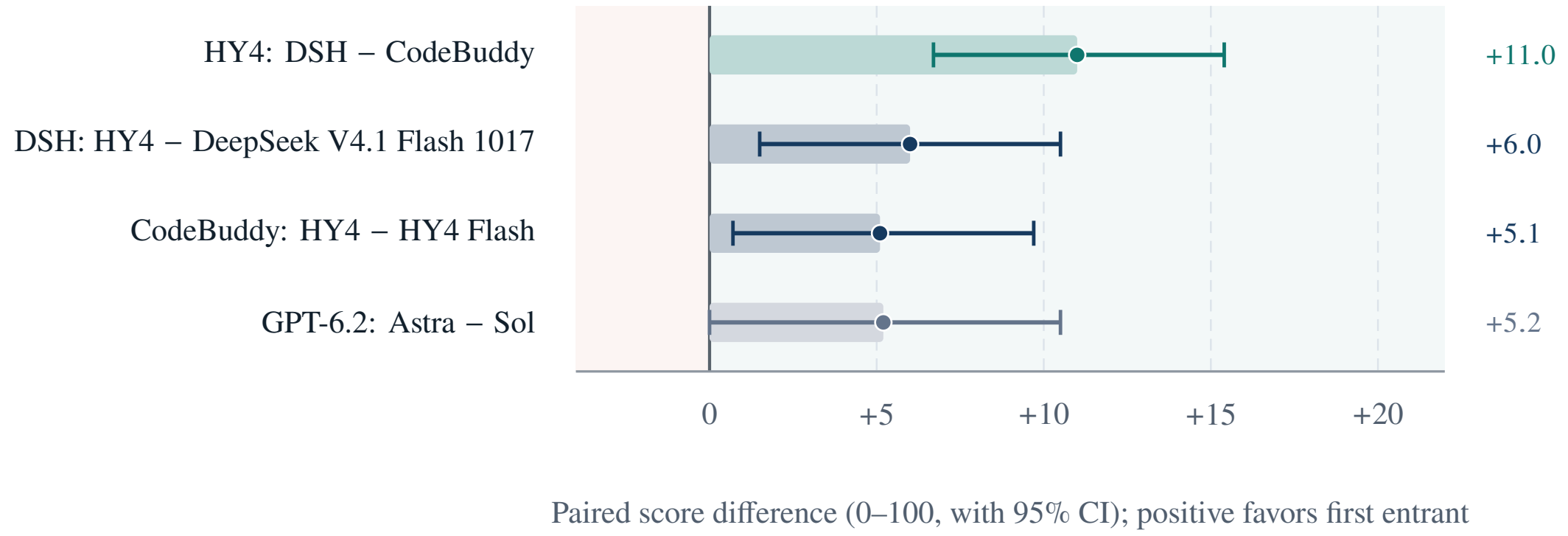
\begin{figure}[H]
\centering
\begin{tikzpicture}[x=3.0mm,y=-8.0mm]
\fill[PKCoral!5](-4,-0.55)rectangle(0,3.55);
\fill[PKTeal!5](0,-0.55)rectangle(22,3.55);
\foreach \t/\lbl in {5/{$+5$},10/{$+10$},15/{$+15$},20/{$+20$}}{\draw[PKRule,densely dashed,line width=0.45pt](\t,-0.55)--(\t,3.55);\node[pknote,anchor=north]at(\t,3.75){\lbl};}
\draw[PKInk!70,line width=0.8pt](0,-0.55)--(0,3.55);
\draw[PKMuted!65,line width=0.6pt](-4,3.55)--(22,3.55);
\node[pknote,anchor=north]at(0,3.75){0};
\node[pknote,anchor=north]at(9,4.85){Paired score difference (0--100, with 95\% CI); positive favors first entrant};
\foreach \i/\name/\v/\lo/\hi/\lbl/\col/\issig in {
 0/{HY4: DSH $-$ CodeBuddy}/11.0/6.7/15.4/{$+11.0$}/PKTeal/1,
 1/{DSH: HY4 $-$ DeepSeek V4.1 Flash 1017}/6.0/1.5/10.5/{$+6.0$}/PKAccent/1,
 2/{CodeBuddy: HY4 $-$ HY4 Flash}/5.1/0.7/9.7/{$+5.1$}/PKAccent/1,
 3/{GPT-6.2: Astra $-$ Sol}/5.2/0.0/10.5/{$+5.2$}/PKSlate/1}{
  \node[anchor=east,font=\footnotesize,text=PKInk]at(-4.8,\i){\name};
  \fill[\col!28,rounded corners=1pt](0,\i-0.22)rectangle(\v,\i+0.22);
  \draw[\col,line width=1.1pt](\lo,\i)--(\hi,\i);
  \draw[\col,line width=0.9pt](\lo,\i-0.14)--(\lo,\i+0.14);
  \draw[\col,line width=0.9pt](\hi,\i-0.14)--(\hi,\i+0.14);
  \ifnum\issig=0\fill[white,draw=\col,line width=1.0pt](\v,\i)circle[radius=2.0pt];\else\fill[\col,draw=white,line width=0.5pt](\v,\i)circle[radius=2.1pt];\fi
  \node[font=\scriptsize\bfseries,text=\col,anchor=west]at(22.8,\i){\lbl};}
\end{tikzpicture}
\caption{\textbf{Swapping the harness on a fixed model (\DSHarnessGap{} points) yields roughly twice the gain of swapping models inside a fixed harness ($+5.1$ to $+6.0$ points).} Controlled paired score differences with 5,000-draw task-bootstrap 95\% CIs (hollow circle denotes CI crossing zero; exact win/loss/tie counts appear in Appendix~\ref{app:stats}, Table~\ref{tab:pairs}).}
\label{fig:contrasts}
\end{figure}

Figure~\ref{fig:contrasts} (with exact paired win/loss counts and intervals in Appendix~\ref{app:stats}, Table~\ref{tab:pairs}) isolates the impact of execution orchestration: \textbf{holding the model fixed at HY4, switching from CodeBuddy to DSH yields \DSHarnessGap{} points (95\% CI \DSHarnessCI{}, \DSHarnessWL{} wins/losses)}, exceeding both model-scale swaps inside CodeBuddy (HY4 vs.\ HY4 Flash: $+5.1$) and cross-family model swaps inside DSH (HY4 vs.\ DeepSeek V4.1 Flash 1017: $+6.0$). Even after excluding all timed-out runs, HY4 + DSH (\DSHyDshOnTime{} on-time mean) still outperforms HY4 + CodeBuddy (\DSHyCbOnTime{}) by 8.6 points, showing that harness quality drives substantive problem-solving gains beyond timeout avoidance and confirming versioned Model + Harness as the necessary evaluation unit.

\FloatBarrier
\subsection{Inter-Agent Agreement, Complementarity, and Frontier Blind Spots}
Motivated by SPU's finding that frontier agents share strong monotone priors and agree far more with one another than with experimental reality (0.52 vs.\ 0.33)~\cite{tu2026spu}, we compute pairwise Pearson correlations of 65-task score vectors across all \DSPairTotal{} configuration pairs. The mean inter-agent correlation reaches \textbf{$r=\DSInterCorr{}$ (range \DSInterCorrRange{})}, showing that current frontier systems stall on largely overlapping scientific barriers.

\begin{figure}[H]
\centering
\includegraphics[width=0.96\linewidth]{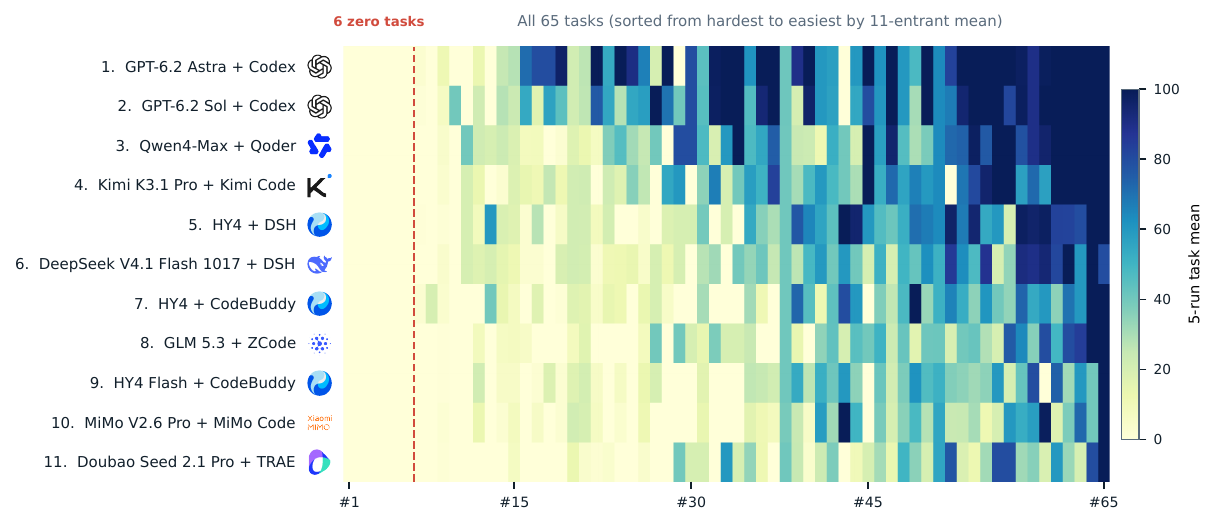}
\caption{\textbf{The full 11 $\times$ 65 score matrix exposes both 6 universally unsolved frontier tasks and distinct cross-system specialization islands.} Five-run mean scores for all 11 configurations across all 65 leaderboard tasks (columns sorted from hardest to easiest by 11-entrant mean).}
\label{fig:task_agent_heatmap}
\end{figure}

The full task--agent matrix (Figure~\ref{fig:task_agent_heatmap}) visualizes both shared blind spots and localized specialization. On the left, \textbf{\DSUnsolvedAll{} tasks score 0.0 across every entrant} (\texttt{R-DAN-12}, \texttt{R-DAN-13}, \texttt{R-SOT-40}, \texttt{R-SOT-41}, \texttt{R-SOT-42}, \texttt{q07}), and pooling all 11 systems passes scientific acceptance on only \textbf{\DSSolvedAny{} / \DSNumTasks{} tasks} (leaving 18 tasks never passed by any configuration); in the middle columns, Tier-2 systems repeatedly form specialization islands that overtake Tier 1 (notably Qwen4-Max + Qoder on open-route retrosynthesis).

\begin{figure}[H]
\centering
\includegraphics[width=0.96\linewidth]{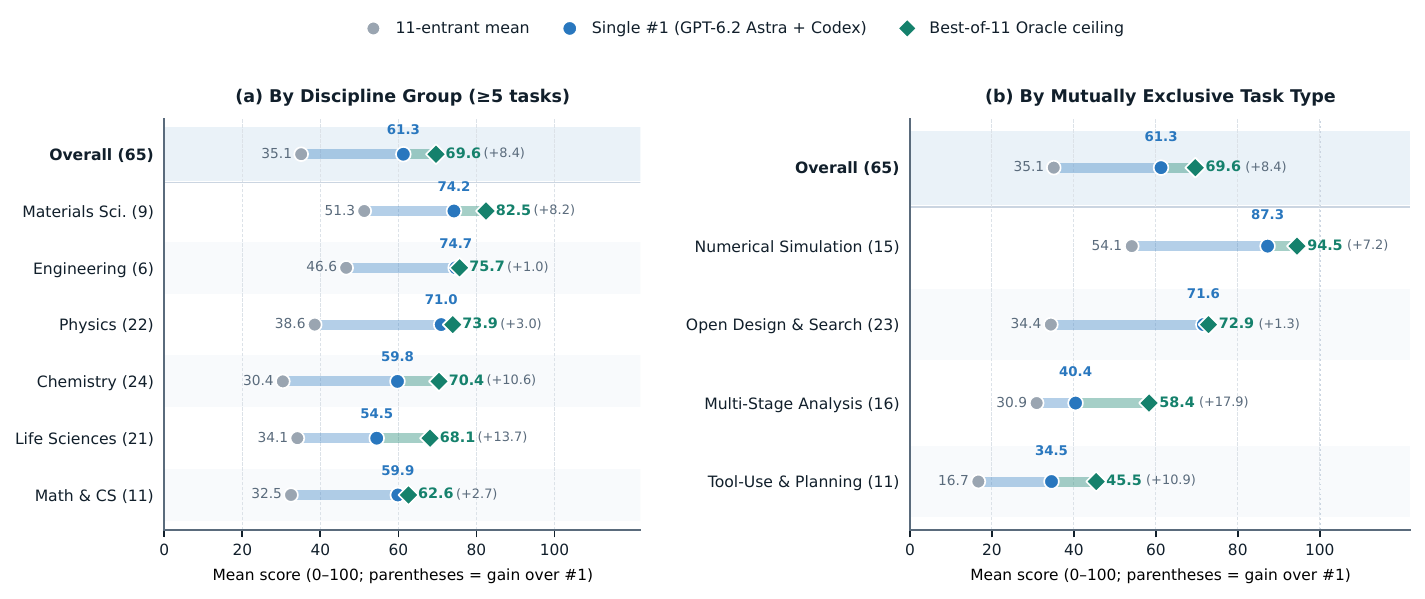}
\caption{\textbf{The Best-of-11 Oracle ceiling (\DSOracleMean{}) exceeds the single \#1 entrant by \DSOracleGain{} points while leaving substantial unsaturated headroom across every domain.} Three-point capability gap comparison across discipline groups (left) and mutually exclusive task types (right): 11-entrant mean (gray circle), single \#1 entrant (blue circle), and Best-of-11 Oracle ceiling (teal diamond).}
\label{fig:difficulty}
\end{figure}

Figure~\ref{fig:difficulty} separates single-system lead from cross-system complementarity: taking the Best-of-11 Oracle score per task lifts the overall mean to \textbf{\DSOracleMean{}}---\textbf{\DSOracleGain{} points above \#1 GPT-6.2 Astra (\DSTopScore{})}---with non-\#1 configurations beating \#1 by $>5$ points on \textbf{\DSNonTopBeats{} tasks} (18.5\%). Complementarity is widest on \textbf{Tool-Use \& Planning} and \textbf{Multi-Stage Analysis}, while Chemistry, Life Sciences, and Math \& CS retain wide unsaturated headroom even under the Oracle ceiling.

\FloatBarrier
\subsection{Time, Token Consumption, Efficiency Frontier, and Failure Modes}
Consistent with SPU's finding that heavier compute budgets do not automatically repair scientific judgment~\cite{tu2026spu}, DeltaScience's execution telemetry confirms the same inefficiency pattern in end-to-end computational research.

\begin{figure}[H]
\centering
\includegraphics[width=0.98\linewidth]{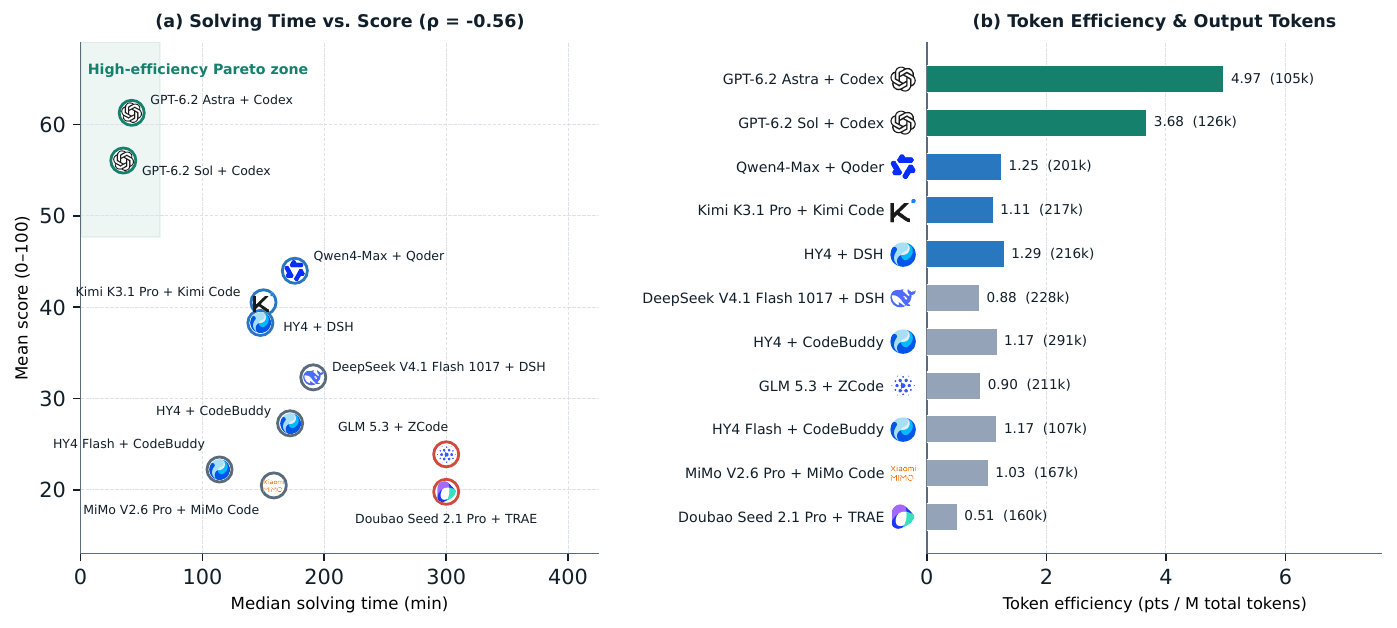}
\caption{\textbf{Median solving time correlates negatively with scientific accuracy ($\rho=\DSRhoTime{}$), with Tier-1 systems occupying a fast, high-efficiency Pareto frontier.} (a) Median solving time vs.\ mean score, with ring colors distinguishing tiers and high-timeout entrants; (b) token efficiency (points per million total tokens) annotated with median output tokens per task.}
\label{fig:pareto}
\end{figure}

Figure~\ref{fig:pareto} exposes a striking divergence between compute spend and scientific accuracy: median solving time correlates negatively with mean score (\textbf{Spearman $\rho=\DSRhoTime{}$}), and output tokens show a weak negative correlation ($\rho=\DSRhoOtok{}$). Tier-1 systems (GPT-6.2 Sol at 35\,min, Astra at 42\,min; 105k--126k median output tokens; 3.68--4.97 points/M total tokens) sit on the Pareto frontier, whereas lower-ranked systems burn 114--300 median minutes and up to 291k output tokens in unproductive retry loops at 0.51--1.29 points/M tokens.

\textbf{Overthinking vs.\ budget-aware convergence: on-time conditional scores.} Comparing overall means against \emph{on-time conditional means} (mean score across non-timeout, non-error trials) separates underlying capability from convergence erosion:
\begin{itemize}
\item \textbf{Latent capability lost to timeouts:} \textbf{GLM 5.3 + ZCode} scores 23.9 overall yet reaches \textbf{\DSGlmOnTime{}} when finishing on time---slightly above \#6 DeepSeek V4.1 Flash 1017 + DSH (\DSDsOnTime{})---losing \textbf{\DSGlmLoss{} points} to its 32.3\% timeout/error rate; \textbf{Doubao Seed 2.1 Pro + TRAE} reaches \textbf{\DSDoubaoOnTime{}} on time (19.8 overall), losing \textbf{\DSDoubaoLoss{} points} to its 37.5\% timeout/error rate.
\item \textbf{Harness control over convergence:} On HY4, DSH finishes in 148\,min (216k output tokens, timeout loss \DSHyDshLoss{} pts; on-time \DSHyDshOnTime{}), whereas CodeBuddy emits 291k output tokens and loses \DSHyCbLoss{} pts to timeouts (on-time \DSHyCbOnTime{}). Even under a generous \DSTimeLimitMed{}-hour median budget, overthinking erodes both user experience and scores, making \textbf{budget-aware convergence} a first-class engineering priority.
\end{itemize}

\begin{table}[htbp]
\centering\small
\caption{\textbf{Open-ended multi-objective search tasks exhibit the highest timeout and error rates, led by \DSTopOutcomeTagEN{} at \DSTopOutcomeRate{}.} Share of combination--task cells with $\ge 1$ recorded timeout, refusal, or harness error by research-task tag (tags covering $\ge 5$ tasks).}
\label{tab:outcomes}
\begin{tabular}{@{}l r r r@{}}
\toprule
Research Task Tag & Tasks & Affected / Total Cells & Share with $\ge 1$ Timeout/Error \\
\midrule
Target selectivity & 5 & 33 / 55 & 60.0\% \\
Molecular design & 6 & 35 / 66 & 53.0\% \\
Configuration optimization & 5 & 28 / 55 & 50.9\% \\
Mechanism analysis & 6 & 24 / 66 & 36.4\% \\
Route evaluation & 12 & 36 / 132 & 27.3\% \\
Synthesis planning & 11 & 29 / 121 & 24.0\% \\
Cell-state analysis & 5 & 11 / 55 & 20.0\% \\
\bottomrule
\end{tabular}

\end{table}

Aligning with the website's failure-mode breakdown (Table~\ref{tab:outcomes}), \textbf{\DSTopOutcomeTagEN{} (\DSTopOutcomeRate{})}, molecular design, and configuration optimization suffer the highest failure incidence. Without crisp stopping signals, open-ended search rapidly drains time budgets whenever physical intuition or stopping heuristics falter.

\FloatBarrier
\subsection{Multi-Dimensional Capability Profiles: Task Types and Discipline Groups}
Table~\ref{tab:typematrix} partitions all 65 tasks into four mutually exclusive task types by deliverable structure and acceptance mechanism.

\begin{table}[htbp]
\centering\scriptsize
\caption{\textbf{Task types reveal sharp specialization: Tool-Use \& Planning flips the top rank while Open Design \& Search opens the widest tier gap.} Five-run mean score matrix across the four mutually exclusive task types (0--100, ordered by overall rank; highest score per column in bold).}
\label{tab:typematrix}
\begin{tabular}{@{}l r r r r r@{}}
\toprule
Model + Harness Configuration & Sim.\ \& Comp. (15) & Open Design (23) & Multi-Stage (16) & Tool-Use (11) & Overall (65) \\
\midrule
\modellogo{gpt}GPT-6.2 Astra + Codex & \textbf{87.3} & \textbf{71.6} & 40.4 & 34.5 & \textbf{61.3} \\
\modellogo{gpt}GPT-6.2 Sol + Codex & 73.3 & 61.1 & \textbf{46.2} & 36.4 & 56.0 \\
\modellogo{qwen}Qwen4-Max + Qoder & 60.6 & 37.2 & 39.6 & \textbf{41.8} & 44.0 \\
\modellogo{kimi}Kimi K3.1 Pro + Kimi Code & 55.8 & 44.6 & 33.1 & 21.8 & 40.5 \\
\modellogo{hy4}HY4 + DSH & 67.2 & 31.8 & 38.1 & 12.7 & 38.3 \\
\modellogo{deepseek}DeepSeek V4.1 Flash 1017 + DSH & 53.4 & 32.9 & 27.8 & 9.1 & 32.3 \\
\modellogo{hy4}HY4 + CodeBuddy & 50.5 & 22.6 & 30.9 & 0.0 & 27.3 \\
\modellogo{glm}GLM 5.3 + ZCode & 42.5 & 19.2 & 20.9 & 12.7 & 23.9 \\
\modellogo{hy4}HY4 Flash + CodeBuddy & 34.2 & 24.9 & 22.4 & 0.0 & 22.2 \\
\modellogo{mimo}MiMo V2.6 Pro + MiMo Code & 41.7 & 17.1 & 19.6 & 0.0 & 20.5 \\
\modellogo{doubao}Doubao Seed 2.1 Pro + TRAE & 29.1 & 15.3 & 21.2 & 14.5 & 19.8 \\
\bottomrule
\end{tabular}

\end{table}

Four distinct capability profiles emerge across execution paradigms:
\begin{itemize}
\item \textbf{Simulation \& Comp.\ (15 tasks):} Strongest category overall, led by GPT-6.2 Astra (87.3) with HY4 + DSH reaching 67.2.
\item \textbf{Open Design \& Search (23 tasks):} Produces the widest tier gap, with GPT-6.2 Astra (71.6) and Sol (61.1) far ahead of Tier 2 (32--45).
\item \textbf{Multi-Stage Analysis (16 tasks):} Compresses the frontier, where GPT-6.2 Sol (46.2) leads GPT-6.2 Astra (40.4) and Qwen4-Max (39.6).
\item \textbf{Tool-Use \& Planning (11 tasks, open-route retrosynthesis):} Flips the leaderboard---\textbf{Qwen4-Max + Qoder (41.8) surpasses GPT-6.2 (34.5--36.4) for first place}, while three harnesses unable to orchestrate the specialized toolchain score 0.0.
\end{itemize}

\begin{table}[htbp]
\centering\small
\caption{\textbf{GPT-6.2 systems lead all seven discipline groups, while Chemistry (30.4), Math \& CS (32.5), and Life Sciences (34.1) record the lowest field averages.} 11-entrant means, top two configurations, and paired top-two 95\% CIs across all 7 discipline groups (* Earth Science has $n=1$ and is listed for reference only).}
\label{tab:groupperf}
\begin{tabularx}{\linewidth}{@{}l r r >{\raggedright\arraybackslash}X >{\raggedright\arraybackslash}X l@{}}
\toprule
Discipline Group & Tasks & 11-Mean & \#1 Configuration & \#2 Configuration & Gap \& 95\% CI \\
\midrule
Chemistry & 24 & 30.4 & GPT-6.2 Astra (59.8) & GPT-6.2 Sol (55.5) & $+4.3$ $[-7.4, +15.7]$ \\
Physics & 22 & 38.6 & GPT-6.2 Astra (71.0) & GPT-6.2 Sol (58.9) & $+12.1$ $[+5.8, +19.5]$ (sep.) \\
Life Sciences & 21 & 34.1 & GPT-6.2 Astra (54.5) & GPT-6.2 Sol (53.0) & $+1.4$ $[-10.9, +13.5]$ \\
Math \& CS & 11 & 32.5 & GPT-6.2 Astra (59.9) & GPT-6.2 Sol (58.4) & $+1.5$ $[-1.8, +5.3]$ \\
Materials Science & 9 & 51.3 & GPT-6.2 Astra (74.2) & GPT-6.2 Sol (67.0) & $+7.3$ $[+0.6, +14.3]$ (sep.) \\
Engineering & 6 & 46.6 & GPT-6.2 Astra (74.7) & GPT-6.2 Sol (73.7) & $+1.0$ $[+0.1, +2.2]$ (sep.) \\
Earth Science * & 1 & 61.2 & GPT-6.2 Astra (100.0) & GPT-6.2 Sol (100.0) & --- \\
\bottomrule
\end{tabularx}

\end{table}

Across the seven discipline groups (Table~\ref{tab:groupperf}), top scores already exceed 71 points in Engineering (74.7), Materials Science (74.2), and Physics (71.0), whereas Chemistry (24 tasks), Math \& CS (11 tasks), and Life Sciences (21 tasks) combine the lowest field averages with the tightest top-two margins.

\begin{figure}[H]
\centering
\includegraphics[width=0.90\linewidth]{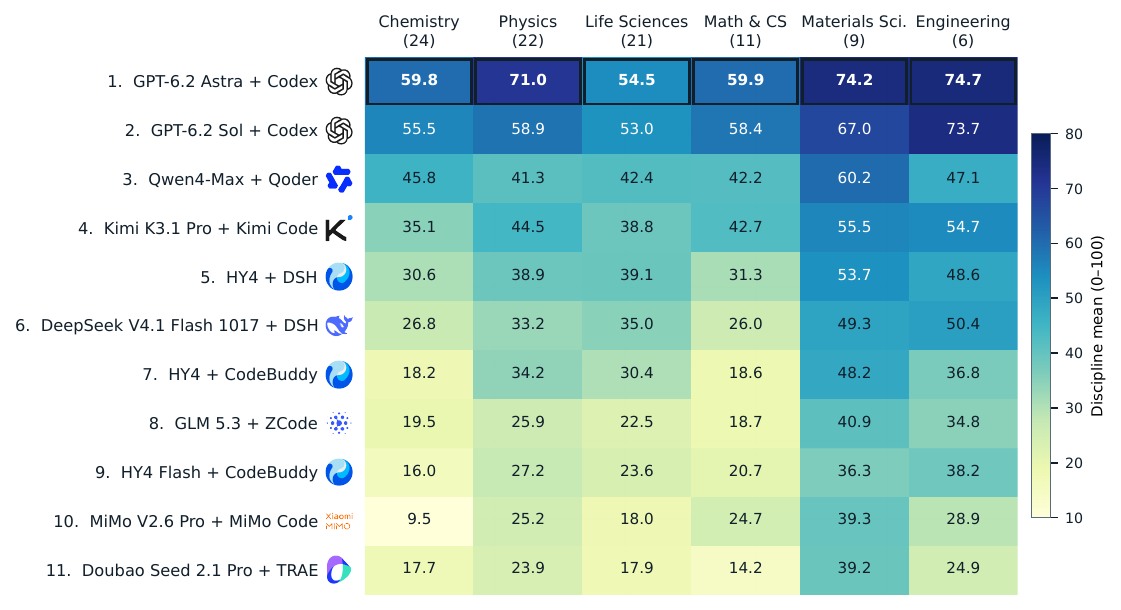}
\caption{\textbf{The 11 $\times$ 6 discipline heatmap highlights Tier-2 specialization, with Qwen4-Max + Qoder and Kimi K3.1 Pro trading third place across domains.} Five-run mean scores across all six comparable discipline groups ($\ge 5$ tasks; dark frames highlight the top score in each discipline).}
\label{fig:discipline_heatmap}
\end{figure}

Figure~\ref{fig:discipline_heatmap} details these disciplinary strengths across all 11 entrants: \textbf{Qwen4-Max + Qoder} claims third place in Chemistry, Life Sciences, and Materials Science, whereas \textbf{Kimi K3.1 Pro + Kimi Code} takes third in Physics, Math \& CS, and Engineering---directly reflecting the \DSVarResid{} task--configuration interaction variance.

\textbf{Task difficulty gradient and high-discrimination tasks.} Across all 65 tasks, per-task means span a wide gradient (min 0.0, Q1 \DSTaskQOne{}, median \textbf{\DSTaskMed{}}, Q3 \DSTaskQThree{}, max \DSTaskMax{}) with only two near-saturated items (\texttt{q04} at 98.2, \texttt{q06} at 87.3). Highest-discrimination tasks by cross-configuration SD are \texttt{R-DAN-16} (42.4), \texttt{R-DAN-17} (42.2), \texttt{z-01} (40.9), \texttt{LKM-0014} (39.1), \texttt{R-DAN-26} (39.1), \texttt{q01} (38.0), and \texttt{QI-0002} (35.8). Section~\ref{sec:cases} moves beneath these aggregate scores to inspect execution traces and microscopic scientific failure modes.
\FloatBarrier

\section{Case Studies and Execution Trace Attribution}\label{sec:cases}

Aggregate scores establish overall rankings and capability specialization, but cannot by themselves explain \emph{why} an agent failed scientific acceptance. Leveraging full execution logs, intermediate states, and deliverables archived in Wenyon, this section moves beneath headline metrics to dissect macro zero-score attribution, four microscopic scientific cases, and the structured diagnostic feedback loop.

\subsection{Macro Zero-Score Trace Attribution: Execution vs.\ Scientific Judgment}
Dissecting all zero-score runs across the inaugural evaluation reveals three primary failure sources:
\begin{enumerate}
\item \textbf{Completed execution failing scientific verification (68.4\%):} Most zero scores do not stem from environment setup or script crashes. Agents reliably install dependencies, invoke solvers, and emit syntactically valid deliverables, yet miss physical conservation laws, numerical convergence criteria, or chemical mechanisms---directly echoing SPU's finding that scientific judgment rather than tool execution is the primary bottleneck~\cite{tu2026spu}.
\item \textbf{Budget-exhausted timeouts (28.2\%):} On open-ended multi-objective search and high-dimensional optimization tasks (such as \texttt{O07}, \texttt{z-02}, \texttt{CWQ-4}, and \texttt{T14}), weaker configurations lack domain-grounded convergence criteria and exhaust time budgets in repetitive retry loops, aligning with Table~\ref{tab:outcomes}.
\item \textbf{Explicit refusals or harness execution errors (3.4\%):} A small fraction of runs terminate via model refusals or harness orchestration failures (e.g., repeated failures on \texttt{LH-1/2/3}).
\end{enumerate}

\subsection{Microscopic Case Studies}
\textbf{Case 1: \texttt{T14} (Materials Science / Open Design)---Microscopic pore blockades and timeout failure.} Task \texttt{T14} requires jointly optimizing 504 discrete trap orientations and 72 continuous processing controls across 504 pores, 192 scenarios, and forward/reverse histories to release mobile species A while retaining bound species B. Whereas the reference design achieves \textbf{99.993\%} worst-pore selective yield (retaining 99.999988\% of B), frontier agents score only 21.4--22.1 or 0.

Microscopic inspection of Pore 104 traces explains the gap: valid-format submissions satisfied global inventory and processing budgets yet left \textbf{7 cooperative local blockades}, raising the A-release activation barrier from \textbf{0.36\,eV} (reference) to \textbf{1.16\,eV} so almost no A escaped within the 70\,s window. Meanwhile, HY4 and GLM 5.3 produced intermediate configurations but timed out before submitting a gradable artifact (score 0), underscoring the failure to feed local physical bottlenecks back into outer search.

\textbf{Case 2: \texttt{QI-0002} (Physics / Quantum Error Correction)---Realistic gate-noise trade-offs.} Motivated by silicon spin-qubit error correction experiments (Takeda et al.~\cite{takeda2022quantum}) and variational QEC (VarQEC~\cite{cao2022varqec}, VGQEC~\cite{shao2024vgqec}), agents design a 5-qubit encoding and 2-ancilla recovery circuit ($\le 256$ CX and 256 rotation gates) protecting an unknown state for 4\,$\mu$s across seven storage noise models. Because encoding and recovery gates introduce gate noise themselves, the verifier scans gate error rates up to $T=10^{-4}$ to measure real hardware-noise tolerance rather than noiseless ideal fidelity.

\textbf{Case 3: \texttt{FUS-0001} (Physics / Fusion Equilibrium)---Local time-slice spikes and multi-run variance.} Grounded in MAST-U discharge \#45425~\cite{pentland2025validation}, per-slice trace inspection of a passing HY4 submission (Figure~\ref{fig:fus}) shows a 95th-percentile boundary error of 33.08\,mm (passing at 0.518) alongside a worst single-slice spike of 62.53\,mm, while its five-run trajectory (Figure~\ref{fig:fiveruns}: three full scores, 64.2, and one timeout) demonstrates why single-run averages mask operational risk.

\textbf{Case 4: Toolchain orchestration vs.\ high-dimensional solvers (\texttt{R-DAN} and \texttt{Science-023}).} In open-route retrosynthesis (\texttt{R-DAN-16} and \texttt{R-DAN-17}, SD 42.2--42.4), execution traces highlight decisive harness orchestration gaps: Qwen4-Max + Qoder (41.8, ranked first in this category) iteratively validates precursor inventories and atom mappings via domain tools to surpass GPT-6.2 (34.5--36.4), whereas three unadapted harnesses score 0.0 across all 11 tasks. In 100-dimensional PDE solving via Deep BSDE (\texttt{Science-023}, building on Han et al.~\cite{han2018deepbsde}), leading configurations converge rapidly and verify parameter sensitivity, whereas weaker entrants diverge under hyperparameter oscillations.

\subsection{From Execution Traces to Retestable Diagnostic Feedback}
Backed by these archived artifacts and traces, the DeltaScience platform publishes synchronized interactive views (leaderboard, outcome breakdowns, discipline rankings, research-direction network, harness contrasts, and public case records) while returning structured diagnostics to model and harness teams across \textbf{scientific knowledge}, \textbf{tool use}, and \textbf{domain judgment}---enabling targeted iteration and same-task verification of true scientific increments ($\delta$).

\section{Related Work}\label{sec:related}

Evaluation of AI agents in science and engineering has shifted rapidly from static QA to executable sandboxes. Table~\ref{tab:related} compares DeltaScience with five representative benchmarks.

\begin{table}[htbp]
\centering\footnotesize
\caption{\textbf{DeltaScience extends container-level execution rigor to a private anti-contamination suite, continuous physical verification, 5-run reliability, and three-layer governance.} Comparison with five representative scientific and agentic benchmarks.}
\label{tab:related}
\begin{tabularx}{\linewidth}{@{}
  >{\raggedright\arraybackslash}p{25mm}
  >{\raggedright\arraybackslash}p{23mm}
  >{\raggedright\arraybackslash}p{21mm}
  >{\raggedright\arraybackslash}p{21mm}
  >{\raggedright\arraybackslash}X@{}}
\toprule
Benchmark & Task Scope & Eval.\ Unit & Scoring & Key Distinctions \\
\midrule
Terminal-Bench~\cite{merrill2026terminalbench} & CLI systems \& software engineering (89 tasks) & Model + Agent Harness & Binary container unit tests & Establishes Harbor sandbox rigor for CLI agents; DeltaScience extends this discipline to multi-domain scientific computation and continuous physical acceptance. \\
\addlinespace[4pt]
Terminal-Bench-Science~\cite{tbscience2026} & Open-source expert computational research tasks (v0.1: 70 tasks across 5 domains) & Model + CLI Harness & Binary 100\%-pass container test suites & Targets authentic scientific computing via open GitHub PRs; DeltaScience complements this with a \textbf{private anti-contamination suite}, continuous $[0,1]$ physical scoring for open-ended optimization, multi-label cross-disciplinary taxonomy (44.6\% cross-group), 5-run reliability, and three-layer governance. \\
\addlinespace[4pt]
AstaBench~\cite{bragg2025astabench} & Literature, code, and data-analysis suites & AI Agent & Multi-metric suite aggregation & Broad scientific skill coverage; DeltaScience focuses on single-step frontline deliverables with five-run pinned-set retesting. \\
\addlinespace[4pt]
PaperBench~\cite{starace2025paperbench} & End-to-end AI paper reproduction & AI Agent & Hierarchical grading rubrics & Targets published AI papers; DeltaScience sources active multi-disciplinary research and industrial problems kept private against contamination. \\
\addlinespace[4pt]
SPU~\cite{tu2026spu} & Hidden experimental outcome prediction (100 public + 16 private) & Model + Harness & Task-adjusted Spearman \& trace audits & Exposes shared agent priors (0.52 vs.\ 0.33) and compute-scaling limits in scientific judgment; DeltaScience complements SPU with executable deliverables, 5-run reliability, and three-role open governance. \\
\addlinespace[4pt]
\textbf{DeltaScience (ours)} & \textbf{\DSNumTasks{} frontline scientific \& industrial steps} & \textbf{Versioned Model + Harness} & \textbf{Isolated continuous metrics + pass line} & \textbf{\DSNumGroups{} groups / \DSNumSubfields{} subfields / 4 task types, private anti-contamination suite, 5-run reliability, three-layer governance, CSTR IDs, and staged public release.} \\
\bottomrule
\end{tabularx}
\end{table}

Overall, DeltaScience synthesizes the Harbor container-isolation paradigm of Terminal-Bench and Terminal-Bench-Science with continuous domain verification, corroborates SPU's findings on shared bottlenecks and compute inefficiency in end-to-end scientific execution, and embeds evaluation inside a private, contamination-resistant ``Playground submission---Bench admission---Bohrium execution---same-task retesting'' ecosystem.
\FloatBarrier

\section{Governance, Future Outlook, and Open Ecosystem}\label{sec:community}

The long-term value of a scientific benchmark depends not only on its inaugural release, but on credible academic governance, a continuous supply of frontline problems, and clear boundaries for open participation and compliant usage. This section presents DeltaScience's three-layer decoupled governance architecture, forward-looking evolution roadmap, and open participation channels alongside usage terms.

\subsection{Three-Layer Decoupled Governance Architecture}\label{sec:governance}
\begin{figure}[htbp]
\centering
\begin{tikzpicture}[
  node distance=2.6mm,
  stage/.style={draw=PKAccent!35,fill=PKPaper,rounded corners=3pt,align=center,
    minimum width=20.2mm,minimum height=14.5mm,inner xsep=2pt,inner ysep=4pt,
    font=\footnotesize,text=PKInk,line width=0.65pt},
  stagefut/.style={stage,draw=PKTeal!65,densely dashed,fill=PKTeal!4},
  stepnum/.style={circle,fill=PKAccent,text=white,font=\tiny\bfseries,inner sep=1.2pt,minimum size=4mm},
  stepnumfut/.style={stepnum,fill=PKTeal},
  flowarr/.style={-{Stealth[length=2mm,width=1.6mm]},draw=PKAccent!75,line width=0.8pt},
  looparr/.style={-{Stealth[length=2mm,width=1.6mm]},draw=PKTeal,densely dashed,line width=0.8pt,rounded corners=4pt}
]
\node[stage](a){\textbf{Real Problems}\\[1pt]{\scriptsize\color{PKMuted}Academia \& R\&D}};
\node[stage,right=of a](b){\textbf{Submission \& Review}\\[1pt]{\scriptsize\color{PKMuted}Playground Portal}};
\node[stage,right=of b](c){\textbf{Pinned Package}\\[1pt]{\scriptsize\color{PKMuted}Bench + CSTR}};
\node[stage,right=of c](d){\textbf{5-Run Eval}\\[1pt]{\scriptsize\color{PKMuted}Bohrium Sandbox}};
\node[stage,right=of d](e){\textbf{Diagnostics}\\[1pt]{\scriptsize\color{PKMuted}Scores \& Gaps}};
\node[stagefut,right=of e](f){\textbf{Retesting}\\[1pt]{\scriptsize\color{PKTeal}Verified $\delta$}};

\foreach \nd/\num in {a/1,b/2,c/3,d/4,e/5}{\node[stepnum]at(\nd.north){\num};}
\node[stepnumfut]at(f.north){6};
\foreach \x/\y in {a/b,b/c,c/d,d/e}{\draw[flowarr](\x)--(\y);}
\draw[looparr](e)--(f);
\draw[looparr]([yshift=2.2mm]e.north)--++(0,7.2mm)-|([yshift=2.2mm]a.north);
\node[fill=white,inner xsep=4pt,inner ysep=1pt,font=\scriptsize,text=PKTeal]
  at([yshift=9.4mm]$(a.north)!0.5!(e.north)$){Bottlenecks guide new tasks $\cdot$ saturated tasks open-sourced};
\draw[looparr](f.south)--++(0,-6.2mm)-|(d.south);
\node[fill=white,inner xsep=4pt,inner ysep=1pt,font=\scriptsize,text=PKTeal]
  at([yshift=-6.2mm]$(d.south)!0.5!(f.south)$){New versions return to the same pinned set};
\draw[PKAccent!75,line width=1.2pt]([yshift=-6.2mm]a.south west)--++(5mm,0)
  node[right=1.5mm,font=\scriptsize,text=PKMuted]{v1.0 pipeline};
\draw[PKTeal,densely dashed,line width=0.9pt]([xshift=25mm,yshift=-6.2mm]a.south west)--++(5mm,0)
  node[right=1.5mm,font=\scriptsize,text=PKTeal]{Continuous loop};
\end{tikzpicture}
\caption{\textbf{Three-layer decoupled governance forms a continuous lifecycle from community submission and isolated evaluation to same-task retesting ($\delta$).} Playground handles rolling submissions and authoring competitions, DeltaScience Benchmark governs admission and versioned releases, and Bohrium executes five-run isolated sandbox evaluations.}
\label{fig:pipeline}
\end{figure}
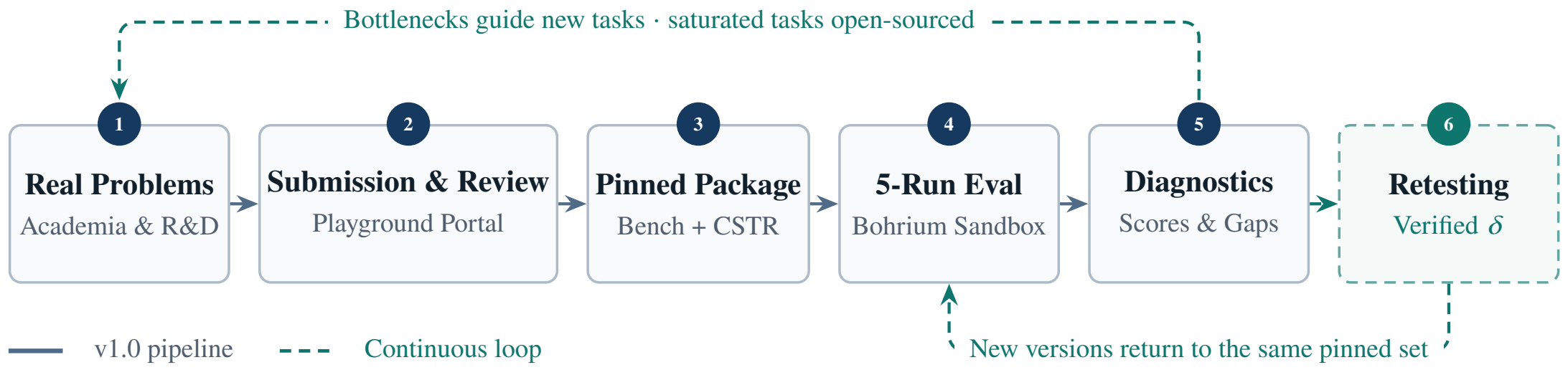

To reconcile academic credibility, sustained community task authoring, and large-scale compute execution, DeltaScience decouples governance into \textbf{three independent operational layers} (Figure~\ref{fig:pipeline}):
\begin{itemize}
\item \textbf{DeltaScience Benchmark (Independent Standards \& Release Authority):} The sole admission and version-release authority, organized by scientific domain. It sets task-admission and scientific acceptance standards, coordinates independent domain-expert review, and publishes pinned monthly releases. DeltaScience Benchmark operates on strictly \textbf{non-profit, academically neutral principles}---tasks are never sold, sublicensed, used for model training, or commercialized, and the authority does not issue monetary prizes directly.
\item \textbf{Playground (Year-Round Community Submission \& Competition Portal):} An open, year-round submission, automated pre-check, and batch peer-review platform hosting themed authoring competitions and contributor incentives. Competition schedules and incentives operate exclusively within Playground, while every admitted task must satisfy DeltaScience Benchmark's unified acceptance criteria.
\item \textbf{Bohrium (Underlying Compute \& Isolated Execution Infrastructure):} Provides unified model routing, isolated execution sandboxes, 10,000+ concurrent container scheduling, and compute resources without participating in admission decisions or hosting competitions.
\end{itemize}

\FloatBarrier

\subsection{Future Outlook and Continuous Evolution Roadmap}\label{sec:roadmap}
Looking ahead, DeltaScience will evolve along four forward-looking directions---expanding disciplinary breadth, bringing in richer real-world industrial data and active research problems, advancing toward long-horizon scientific workflows, and sustaining an open academic ecosystem:

\begin{itemize}
\item \textbf{Expanding disciplinary breadth and problem depth:} Building on v1.0's \DSNumGroups{} discipline groups and \DSNumSubfields{} subfields, we will systematically grow both domain coverage and difficulty headroom---prioritizing currently sparse or emerging frontiers (including Earth and climate sciences, translational medicine, energy and environmental engineering, and microelectronics/advanced manufacturing) while targeting exposed bottlenecks such as multi-objective open design, high-dimensional convergence, and specialized toolchain orchestration.
\item \textbf{Integrating richer industrial data and active frontier research problems:} Under a dual-track \textbf{pinned-set retesting + periodic independent new-task expansion} model (new batches are evaluated and published independently without mixing into historical suites), we will deepen partnerships with pharmaceutical, materials, energy, and engineering R\&D teams alongside frontline academic laboratories. Under strict anonymization and IP compliance, we will incorporate more real-world experimental datasets, production R\&D constraints, and active research problems, extending evaluation from single computational steps toward multi-stage scientific investigations.
\item \textbf{Long-horizon research loops and trajectory-level scientific diagnostics:} Looking toward next-generation autonomous scientific agents, we plan to explore multimodal experimental data interpretation, simulation--experiment feedback loops, and longer-horizon research campaigns. Leveraging Wenyon-archived execution traces, we will also adopt SPU-style trajectory diagnostics~\cite{tu2026spu} to separate tool-execution errors, computational proxy traps, and core scientific-judgment failures---providing actionable signals for scientific pretraining and post-training.
\item \textbf{Academic credit and staged open-sourcing:} Every admitted task receives a persistent \textbf{CSTR (Common Science and Technology Resource)} identifier and collective institutional credit. Active suites remain private to prevent contamination; once frontier systems saturate a subset of tasks, those tasks are systematically graduated and open-sourced to communities such as DeepModeling for open research and reproducibility.
\end{itemize}

\subsection{Open Participation and Submission Channels}\label{sec:channels}
To enable seamless participation across the scientific and engineering community, DeltaScience maintains three year-round open channels (supporting platforms and competition links appear in the footnotes below and Appendix~\ref{app:sources}):
\begin{itemize}
\item \textbf{Academic \& industrial task authors:} Submit frontline scientific and R\&D problems through the year-round Playground Authoring Competition\footnote{Playground Authoring Competition and year-round submission portal: \url{https://play.bohrium.com/competition/science-pilot}} and Agentic Science Challenge. Tasks passing automated pre-checks and independent DeltaScience Benchmark domain review receive collective authorship, persistent CSTR identifiers, citable records, and competition incentives.
\item \textbf{Model \& harness teams:} Submit new versioned Model + Harness configurations via the DeltaScience official platform\footnote{DeltaScience official evaluation platform and leaderboard: \url{https://deltascience.bohrium.com}} to receive per-task diagnostics across scientific knowledge, tool use, and domain judgment, and schedule same-task retesting to measure verified $\delta$.
\item \textbf{Open-source \& academic community:} Explore public leaderboards, multi-dimensional analyses, public case records (T14, QI-0002, FUS-0001), and open-source benchmark packages on GitHub\footnote{DeltaScience open-source repository (public exemplar task packages and benchmark documentation): \url{https://github.com/deepmodeling/DeltaScience}}, and join reproducibility, tool-ecosystem building (e.g., DeepModeling), and domain peer review around graduated public tasks.
\end{itemize}

\textbf{Usage policy and leaderboard interpretation.} DeltaScience benchmark tasks---including public example cases, problem specifications, reference solutions, and verifier scripts---are made available strictly for non-profit scientific evaluation, academic citation, and community reference. No organization or individual may \textbf{use any benchmark task or its paraphrases, translations, or derivatives to train, fine-tune, or distill any model, or to generate synthetic training data}; bulk scraping, commercial sale, paid question-bank redistribution, or other for-profit exploitation is strictly prohibited. Leaderboard scores reflect finite-trial evaluation outcomes for a specific model, harness, parameter configuration, runtime environment, and timestamp; given the stochasticity of large models, results may vary across runs or configurations and do not constitute commercial endorsement.

\section{Conclusion}\label{sec:conclusion}

DeltaScience launches with a clear, disciplined mission: \textbf{connecting frontier scientific discovery with real-world industrial R\&D, and testing AI scientific capability on problems that truly matter.} Positioned as \textbf{an open evaluation platform for large models and AI agents in real scientific scenarios}, DeltaScience bridges real academic and industrial bottlenecks directly to AI agents through a 10,000+ sandbox substrate (Wenyon, Flywheel, three-layer isolated sandboxes, and EvalWheel) and a three-layer decoupled governance ecosystem (DeltaScience Benchmark for independent standards and versioned releases, Playground for year-round community submissions and authoring competitions, and Bohrium for isolated compute infrastructure), anchoring every score to immutable artifacts and execution traces so that capability progress becomes a verifiable scientific increment ($\delta$) driving the continuous evolution of both the benchmark and AI scientific capability.

Across \DSNumRuns{} independent sandbox runs spanning \DSNumTasks{} v1.0 tasks and \DSNumCombos{} frontier configurations, six clear empirical signals emerge: (1) task difficulty explains \DSVarTask{} of score variance, underscoring the value of a broad, cross-disciplinary task suite (median \DSTaskMed{}, \DSCrossPct{} cross-group); (2) swapping the harness on HY4 (\DSHarnessGap{} points, 95\% CI \DSHarnessCI{}) outweighs swapping models within a fixed harness, confirming Model + Harness as the necessary evaluation unit; (3) multi-run reliability remains a primary bottleneck (\DSNeverFullPct{} of cells never pass scientific acceptance across five trials, and even \#1 drops over 10 points from pass@5 to pass\textsuperscript{5}); (4) inter-agent correlations average \DSInterCorr{}, yet the Best-of-11 Oracle (\DSOracleMean{}) reveals \DSOracleGain{} points of complementarity alongside \DSUnsolvedAll{} universally unsolved tasks; (5) longer runtime correlates negatively with score ($\rho=\DSRhoTime{}$), confirming that extra compute cannot substitute for scientific judgment; and (6) capabilities specialize strongly by task type, with Qwen4-Max + Qoder (41.8) surpassing GPT-6.2 (34.5--36.4) on Tool-Use \& Planning.

Real problems. Scientific rigor. Turning AI progress into scientific progress. We invite academic groups, industrial R\&D teams, model and harness developers, and the open-source community to join DeltaScience through the open channels in Section~\ref{sec:channels} and bring the hardest problems in real science into open evaluation.

\section*{Contributors}\label{sec:contributors}
\addcontentsline{toc}{section}{Contributors}

DeltaScience is a collaborative multi-institution effort. Contributors are organized by functional role below, with members contributing across multiple workstreams listed in each applicable group; superscripts denote institutional affiliations.

\begin{description}[leftmargin=1.5em,style=nextline,itemsep=7pt]
\item[\textbf{Scientific Task Authors \& Verifiers}]
\textit{Frontline academic and industrial task authoring, scientific acceptance verifiers, and domain review.}\\[2pt]
\textbf{Xinyu~Shen}\textsuperscript{2},
\textbf{Chenyu~Xie}\textsuperscript{2},
\textbf{Shuang~Wu}\textsuperscript{2},
\textbf{Wen~Xu}\textsuperscript{2},
\textbf{Mingjun~Xu}\textsuperscript{2},
\textbf{Zhirui~Lai}\textsuperscript{2,5},
\textbf{Xiaoyang~Liu}\textsuperscript{2},
\textbf{Jing~Su}\textsuperscript{2,4},
\textbf{Jia~Guo}\textsuperscript{2},
\textbf{Yuxuan~Ma}\textsuperscript{1,4},
\textbf{Junjie~Wang}\textsuperscript{3},
\textbf{Ruoshui~Xu}\textsuperscript{1,6},
\textbf{Changxi~Chen}\textsuperscript{1},
\textbf{Weiqi~Chen}\textsuperscript{2},
\textbf{Shifan~Chen}\textsuperscript{2,4},
\textbf{Hang~Lin}\textsuperscript{2,9},
\textbf{Yifeng~Guan}\textsuperscript{1,4},
\textbf{Xinjie~Zhou}\textsuperscript{1,4},
\textbf{Qi~Liu}\textsuperscript{2},
\textbf{Qingguo~Zhou}\textsuperscript{1}

\item[\textbf{Evaluation Infrastructure, Platform \& Analysis}]
\textit{Large-scale isolated sandbox substrate, evaluation execution, empirical analysis, model hosting, and platform operations.}\\[2pt]
\textbf{Mingjun~Xu}\textsuperscript{2},
\textbf{Shigang~Ou}\textsuperscript{2,7},
\textbf{Kai~Cao}\textsuperscript{2},
\textbf{Yi~Wang}\textsuperscript{2},
\textbf{Ye~Li}\textsuperscript{2},
\textbf{Yongkang~Zhai}\textsuperscript{2},
\textbf{Jiaxing~Zhang}\textsuperscript{2},
\mbox{\textbf{Jintan~Lu}\textsuperscript{2}},
\textbf{Wenbo~Shen}\textsuperscript{2}

\item[\textbf{Advisors \& Leadership}]
\textit{Strategic direction, disciplinary guidance, program leadership, and multi-institution governance.}\\[2pt]
\textbf{Xinyu~Li}\textsuperscript{1},
\textbf{Qingguo~Zhou}\textsuperscript{1},
\textbf{Yuan~Huang}\textsuperscript{2},
\textbf{Tianhan~Zhang}\textsuperscript{2,8},
\textbf{Ruoxue~Liao}\textsuperscript{2},
\textbf{Linfeng~Zhang}\textsuperscript{1,2}
\end{description}

\vspace{4pt}
\noindent{\small\color{PKMuted}\textbf{Affiliations:}
\textsuperscript{1}AI for Science Institute, Beijing \quad
\textsuperscript{2}DP Technology \quad
\textsuperscript{3}Peking University \quad
\textsuperscript{4}Zhejiang University \quad
\textsuperscript{5}University of Tokyo \quad
\textsuperscript{6}University of Cambridge \quad
\textsuperscript{7}Institute of Physics, Chinese Academy of Sciences \quad
\textsuperscript{8}Beihang University \quad
\textsuperscript{9}Tongji University}

\bibliographystyle{plain}
\bibliography{references}

\begin{thebibliography}{10}

\bibitem{bragg2025astabench}
Jonathan Bragg, Mike D'Arcy, Nishant Balepur, et~al.
\newblock {AstaBench}: Rigorous benchmarking of {AI} agents with a scientific research suite.
\newblock arXiv:2510.21652, 2025.
\newblock Revised April 2026.

\bibitem{cao2022varqec}
Chenfeng Cao, Chao Zhang, Zipeng Wu, Markus Grassl, and Bei Zeng.
\newblock Quantum variational learning for quantum error-correcting codes.
\newblock arXiv:2204.03560, 2022.

\bibitem{chen2021codex}
Mark Chen, Jerry Tworek, Heewoo Jun, et~al.
\newblock Evaluating large language models trained on code.
\newblock arXiv:2107.03374, 2021.

\bibitem{han2018deepbsde}
Jiequn Han, Arnulf Jentzen, and Weinan E.
\newblock Solving high-dimensional partial differential equations using deep learning.
\newblock {\em Proceedings of the National Academy of Sciences}, 115(34):8505--8510, 2018.

\bibitem{merrill2026terminalbench}
Mike~A. Merrill, Alexander~G. Shaw, Nicholas Carlini, et~al.
\newblock {Terminal-Bench}: Benchmarking agents on hard, realistic tasks in command line interfaces.
\newblock arXiv:2601.11868, 2026.

\bibitem{pentland2025validation}
K.~Pentland, N.~C. Amorisco, O.~El-Zobaidi, et~al.
\newblock Validation of the static forward {Grad--Shafranov} equilibrium solvers in {FreeGSNKE} and {Fiesta} using {EFIT++} reconstructions from {MAST-U}.
\newblock {\em Physica Scripta}, 100(2):025608, 2025.

\bibitem{shao2024vgqec}
Yuxuan Shao, Fuchuan Wei, Song Cheng, and Zhengwei Liu.
\newblock Variational graph-based quantum error correction ({VGQEC}).
\newblock arXiv:2410.02608, 2024.

\bibitem{starace2025paperbench}
Giulio Starace, Oliver Jaffe, Dane Sherburn, et~al.
\newblock {PaperBench}: Evaluating {AI}'s ability to replicate {AI} research.
\newblock arXiv:2504.01848, 2025.

\bibitem{takeda2022quantum}
Kenta Takeda, Akito Noiri, Takashi Nakajima, Takashi Kobayashi, and Seigo Tarucha.
\newblock Quantum error correction with silicon spin qubits.
\newblock {\em Nature}, 608:682--686, 2022.

\bibitem{tbscience2026}
{Terminal-Bench-Science Team} and {Harbor Framework}.
\newblock {Terminal-Bench-Science}: Expert-curated scientific computing tasks for terminal agents.
\newblock GitHub Repository, v0.1.0, 2026.

\bibitem{tu2026spu}
Zengji Tu, Xiang Liu, Jiarui Sun, et~al.
\newblock {Science Prediction Universe}.
\newblock arXiv:2611.15049, 2026.

\bibitem{yao2024taubench}
Shunyu Yao, Noah Shinn, Pedram Razavi, and Karthik Narasimhan.
\newblock {$\tau$-bench}: A benchmark for tool-agent-user interaction in real-world domains.
\newblock arXiv:2406.12045, 2024.

\end{thebibliography}

\clearpage
\appendix
\startcontents[appendices]
\section*{Appendix Table of Contents}
\printcontents[appendices]{}{1}{\small}
\vspace{8pt}{\color{PKRule}\hrule height 0.5pt}\vspace{6pt}
\section{Statistical Methodology}\label{app:stats}
\textbf{Run-to-run variance.} Let $v_{ia}$ be the sample variance across the five trials of configuration $a$ on task $i$. Under independent trials, the run-level variance of the 0--100 overall mean is estimated by
\[
\widehat{\mathrm{Var}}(S_a)=\frac{100^2}{N^2}\sum_{i=1}^{N}\frac{v_{ia}}{5},\qquad N=\DSNumTasks{}.
\]

\textbf{Task-composition sensitivity.} We draw 5,000 paired task-bootstrap resamples of size \DSNumTasks{} with replacement (fixed seed), retaining all configurations on each sampled task, and take the 2.5th and 97.5th percentiles as 95\% confidence intervals.

\textbf{Multi-run reliability (pass@$k$ and pass\textsuperscript{5}).} If $c_{ia}$ of five trials pass a task's binary (0/1) scientific acceptance threshold ($p_{iar}=1$), the unbiased estimator for passing at least once in $k$ attempts is~\cite{chen2021codex}
\[
\widehat{p}_{ia}(k)=1-\frac{\binom{5-c_{ia}}{k}}{\binom{5}{k}},\qquad k=1,\ldots,5,
\]
with numerator 0 when $5-c_{ia}<k$, averaged across all \DSNumTasks{} tasks. Following $\tau$-bench~\cite{yao2024taubench}, we further define \textbf{pass\textsuperscript{5}} as the fraction of tasks passing scientific acceptance in all five trials ($c_{ia}=5$).

\textbf{Pairwise dominance and correlations.} Paired contrasts compute task-wise differences of five-run means and 5,000-draw bootstrap 95\% CIs across all \DSPairTotal{} configuration pairs (\DSPairSep{} separated). Table~\ref{tab:pairs} provides the full paired counts and confidence intervals backing Figure~\ref{fig:contrasts} in the main text. Inter-agent agreement is the mean pairwise Pearson correlation across 65-task score vectors; runtime--accuracy association uses Spearman rank correlation.

\begin{table}[htbp]
\centering\footnotesize\setlength{\tabcolsep}{4.5pt}
\caption{\textbf{Controlled harness swaps outperform fixed-harness model swaps in both score gain and win/loss margin.} Exact paired score contrasts (first minus second), 5,000-draw task-bootstrap 95\% CIs, and task-level win/loss/tie counts backing Figure~\ref{fig:contrasts}.}
\label{tab:pairs}
\begin{tabular}{@{}lrrrrl@{}}
\toprule
Paired Contrast & Shared & Gap & 95\% CI & W / L & Primary Signal \\
\midrule
HY4: DSH $-$ CodeBuddy & 65 & $+11.0$ & $[+6.7, +15.4]$ & 42 / 10 & Harness effect (sig.) \\
DSH: HY4 $-$ DeepSeek V4.1 Flash 1017 & 65 & $+6.0$ & $[+1.5, +10.5]$ & 36 / 16 & Model swap in DSH (sig.) \\
CodeBuddy: HY4 $-$ HY4 Flash & 65 & $+5.1$ & $[+0.7, +9.7]$ & 32 / 13 & Model tier in CodeBuddy (sig.) \\
GPT-6.2: Astra $-$ Sol & 65 & $+5.2$ & $[+0.0, +10.5]$ & 28 / 10 & Same-family variant \\
\bottomrule
\end{tabular}

\end{table}

\section{Resource Accounting}\label{app:resource}
Total tokens sum returned input and output tokens (cached tokens inside input, reasoning tokens inside output, without double counting). Solving time records median agent-solving minutes, excluding scheduling, queuing, and verifier runtime. Token efficiency divides mean score by mean total tokens (millions).

\section{Platform Links and Reproducibility}\label{app:sources}
\begin{itemize}
\item \textbf{Platform Website \& Public Case Studies:} \url{https://deltascience.bohrium.com} (leaderboard, discipline-group rankings, research-direction network, Harness contrasts, and public case records for T14, QI-0002, and FUS-0001).
\item \textbf{Open-Source Repository:} \url{https://github.com/deepmodeling/DeltaScience} (open-source exemplar task packages, task format documentation, and community updates).
\item \textbf{Year-Round Authoring Competition:} \url{https://play.bohrium.com/competition/science-pilot} (Playground open task-submission, pre-check, and batch peer-review portal).
\item \textbf{Official Contact:} \href{mailto:deltascience@aisi.ac.cn}{\texttt{deltascience@aisi.ac.cn}} (model onboarding, task co-authoring, compliance inquiries, and academic partnerships).
\item \textbf{DeltaScience v1.0 Metadata:} \DSNumTasks{} tasks across \DSNumGroups{} discipline groups, \DSNumSubfields{} subfields, 4 task types, and \DSTimeLimitMin{}--\DSTimeLimitMax{}\,h per-task time limits, co-built by 33 contributors across 9 institutions.
\item \textbf{Data-Driven Regeneration:} All quantitative macros, tables, and vector figures in this report are auto-generated by \path{scripts/generate_report_data.py} and \path{scripts/generate_figures.py} for one-command updates when subsequent releases arrive.
\end{itemize}

\section{Sanitized Directory of All 65 DeltaScience v1.0 Tasks}\label{app:tasklist}
To prevent data contamination, problem statements, reference solutions, and verifier scripts remain private. Table~\ref{tab:tasklist} lists sanitized metadata---including task ID, CSTR identifier suffix, discipline groups, subfields, task type, per-task time limit, and contributing institutions---for all \DSNumTasks{} v1.0 leaderboard tasks.

\begingroup\scriptsize\setlength{\tabcolsep}{3.5pt}
\begin{longtable}{@{}r l l >{\raggedright\arraybackslash}p{17mm} >{\raggedright\arraybackslash}p{25mm} >{\raggedright\arraybackslash}p{18mm} r >{\raggedright\arraybackslash}p{20mm}@{}}
\caption{Sanitized metadata directory, per-task evaluation time limits, and CSTR identifiers of all 65 DeltaScience v1.0 leaderboard tasks (CSTR prefix \texttt{31254.49.DELTASCIENCE.} omitted in-cell for compactness, with each entry linked to its resolver URL; problem statements and grading details omitted).}\label{tab:tasklist}\\
\toprule
\# & Task ID & CSTR Suffix & Discipline Group(s) & Subfield(s) & Task Type & Limit & Institution(s) \\
\midrule
\endfirsthead
\toprule
\# & Task ID & CSTR Suffix & Discipline Group(s) & Subfield(s) & Task Type & Limit & Institution(s) \\
\midrule
\endhead
\bottomrule
\endfoot
1 & \texttt{C-01} & \href{https://cstr.cn/31254.49.DELTASCIENCE.V1.C.01}{\texttt{V1.C.01}} & Chemistry, Life Sciences & Medicinal Chem., Cheminformatics, Structural Bio. & Open Design \& Search & 5h & DP Technology, AI for Science Institute, Beijing \\
2 & \texttt{C-02} & \href{https://cstr.cn/31254.49.DELTASCIENCE.V1.C.02}{\texttt{V1.C.02}} & Chemistry, Life Sciences & Medicinal Chem., Cheminformatics, Structural Bio. & Open Design \& Search & 5h & DP Technology, AI for Science Institute, Beijing \\
3 & \texttt{C-03} & \href{https://cstr.cn/31254.49.DELTASCIENCE.V1.C.03}{\texttt{V1.C.03}} & Chemistry, Life Sciences & Medicinal Chem., Cheminformatics, Structural Bio. & Open Design \& Search & 5h & DP Technology, AI for Science Institute, Beijing \\
4 & \texttt{C-04} & \href{https://cstr.cn/31254.49.DELTASCIENCE.V1.C.04}{\texttt{V1.C.04}} & Chemistry, Life Sciences & Medicinal Chem., Cheminformatics, Structural Bio. & Open Design \& Search & 5h & DP Technology, AI for Science Institute, Beijing \\
5 & \texttt{C-05} & \href{https://cstr.cn/31254.49.DELTASCIENCE.V1.C.05}{\texttt{V1.C.05}} & Chemistry, Life Sciences & Medicinal Chem., Chemical Bio., Structural Bio. & Open Design \& Search & 5h & DP Technology, AI for Science Institute, Beijing \\
6 & \texttt{C-06} & \href{https://cstr.cn/31254.49.DELTASCIENCE.V1.C.06}{\texttt{V1.C.06}} & Chemistry, Life Sciences & Medicinal Chem., Chemical Bio., Structural Bio. & Open Design \& Search & 5h & DP Technology, AI for Science Institute, Beijing \\
7 & \texttt{R-DAN-12} & \href{https://cstr.cn/31254.49.DELTASCIENCE.V1.R.DAN.12}{\texttt{V1.R.DAN.12}} & Chemistry & Organic Chem., Medicinal Chem. & Tool-Use \& Planning & 8h & DP Technology, University of Tokyo \\
8 & \texttt{R-DAN-13} & \href{https://cstr.cn/31254.49.DELTASCIENCE.V1.R.DAN.13}{\texttt{V1.R.DAN.13}} & Chemistry & Organic Chem., Medicinal Chem. & Tool-Use \& Planning & 8h & DP Technology, University of Tokyo \\
9 & \texttt{R-DAN-14} & \href{https://cstr.cn/31254.49.DELTASCIENCE.V1.R.DAN.14}{\texttt{V1.R.DAN.14}} & Chemistry & Organic Chem., Medicinal Chem. & Tool-Use \& Planning & 8h & DP Technology, University of Tokyo, Zhejiang University \\
10 & \texttt{R-DAN-15} & \href{https://cstr.cn/31254.49.DELTASCIENCE.V1.R.DAN.15}{\texttt{V1.R.DAN.15}} & Chemistry & Organic Chem., Medicinal Chem. & Tool-Use \& Planning & 8h & DP Technology, University of Tokyo, Zhejiang University \\
11 & \texttt{R-DAN-16} & \href{https://cstr.cn/31254.49.DELTASCIENCE.V1.R.DAN.16}{\texttt{V1.R.DAN.16}} & Chemistry & Organic Chem., Medicinal Chem. & Tool-Use \& Planning & 8h & University of Tokyo, DP Technology \\
12 & \texttt{R-DAN-17} & \href{https://cstr.cn/31254.49.DELTASCIENCE.V1.R.DAN.17}{\texttt{V1.R.DAN.17}} & Chemistry & Organic Chem., Medicinal Chem. & Tool-Use \& Planning & 8h & DP Technology, University of Tokyo \\
13 & \texttt{R-DAN-18} & \href{https://cstr.cn/31254.49.DELTASCIENCE.V1.R.DAN.18}{\texttt{V1.R.DAN.18}} & Chemistry & Organic Chem., Medicinal Chem. & Tool-Use \& Planning & 8h & DP Technology, University of Tokyo \\
14 & \texttt{R-DAN-26} & \href{https://cstr.cn/31254.49.DELTASCIENCE.V1.R.DAN.26}{\texttt{V1.R.DAN.26}} & Chemistry & Organic Chem., Medicinal Chem. & Tool-Use \& Planning & 8h & DP Technology, University of Tokyo \\
15 & \texttt{R-SOT-40} & \href{https://cstr.cn/31254.49.DELTASCIENCE.V1.R.SOT.40}{\texttt{V1.R.SOT.40}} & Chemistry & Organic Chem., Medicinal Chem. & Tool-Use \& Planning & 8h & DP Technology, University of Tokyo \\
16 & \texttt{R-SOT-41} & \href{https://cstr.cn/31254.49.DELTASCIENCE.V1.R.SOT.41}{\texttt{V1.R.SOT.41}} & Chemistry & Organic Chem., Medicinal Chem. & Tool-Use \& Planning & 8h & DP Technology, University of Tokyo, Peking University \\
17 & \texttt{R-SOT-42} & \href{https://cstr.cn/31254.49.DELTASCIENCE.V1.R.SOT.42}{\texttt{V1.R.SOT.42}} & Chemistry & Organic Chem., Medicinal Chem. & Tool-Use \& Planning & 8h & DP Technology, University of Tokyo \\
18 & \texttt{O02} & \href{https://cstr.cn/31254.49.DELTASCIENCE.V1.O02}{\texttt{V1.O02}} & Life Sciences & Omics, Immunology \& Inflamm. & Multi-Stage Analysis & 4h & DP Technology, Zhejiang University \\
19 & \texttt{O03} & \href{https://cstr.cn/31254.49.DELTASCIENCE.V1.O03}{\texttt{V1.O03}} & Life Sciences & Omics, Immunology \& Inflamm. & Multi-Stage Analysis & 8h & AI for Science Institute, Beijing, Zhejiang University \\
20 & \texttt{O04} & \href{https://cstr.cn/31254.49.DELTASCIENCE.V1.O04}{\texttt{V1.O04}} & Life Sciences & Omics, Cancer Biology & Multi-Stage Analysis & 8h & AI for Science Institute, Beijing, University of Cambridge \\
21 & \texttt{O05} & \href{https://cstr.cn/31254.49.DELTASCIENCE.V1.O05}{\texttt{V1.O05}} & Life Sciences & Omics, Immunology \& Inflamm. & Multi-Stage Analysis & 8h & AI for Science Institute, Beijing, Zhejiang University \\
22 & \texttt{O06} & \href{https://cstr.cn/31254.49.DELTASCIENCE.V1.O06}{\texttt{V1.O06}} & Life Sciences & Omics, Cancer Biology & Multi-Stage Analysis & 8h & AI for Science Institute, Beijing, Zhejiang University \\
23 & \texttt{O07} & \href{https://cstr.cn/31254.49.DELTASCIENCE.V1.O07}{\texttt{V1.O07}} & Life Sciences & Omics, Stem Cell \& Aging & Multi-Stage Analysis & 16h & DP Technology, Zhejiang University \\
24 & \texttt{O08} & \href{https://cstr.cn/31254.49.DELTASCIENCE.V1.O08}{\texttt{V1.O08}} & Life Sciences & Omics, Stem Cell \& Aging & Multi-Stage Analysis & 8h & DP Technology, Zhejiang University \\
25 & \texttt{O10} & \href{https://cstr.cn/31254.49.DELTASCIENCE.V1.O10}{\texttt{V1.O10}} & Life Sciences & Omics, Cancer Biology & Multi-Stage Analysis & 8h & AI for Science Institute, Beijing, University of Cambridge \\
26 & \texttt{QI-0001} & \href{https://cstr.cn/31254.49.DELTASCIENCE.V1.QI.0001}{\texttt{V1.QI.0001}} & Physics, Math \& CS & Quantum Comp., Control Theory & Open Design \& Search & 3h & DP Technology \\
27 & \texttt{QI-0002} & \href{https://cstr.cn/31254.49.DELTASCIENCE.V1.QI.0002}{\texttt{V1.QI.0002}} & Physics, Math \& CS & Quantum Comp., Coding Theory & Open Design \& Search & 12h & DP Technology \\
28 & \texttt{QI-0003} & \href{https://cstr.cn/31254.49.DELTASCIENCE.V1.QI.0003}{\texttt{V1.QI.0003}} & Physics, Engineering & Quantum Comp., Eng. Design & Open Design \& Search & 4h & DP Technology \\
29 & \texttt{QI-0005} & \href{https://cstr.cn/31254.49.DELTASCIENCE.V1.QI.0005}{\texttt{V1.QI.0005}} & Physics, Math \& CS & Quantum Comp., Algebra \& Comb., Algorithm Design & Open Design \& Search & 12h & DP Technology \\
30 & \texttt{QI-0006} & \href{https://cstr.cn/31254.49.DELTASCIENCE.V1.QI.0006}{\texttt{V1.QI.0006}} & Physics, Math \& CS & Quantum Comp., Algebra \& Comb., Algorithm Design & Open Design \& Search & 5h & DP Technology \\
31 & \texttt{QI-0007} & \href{https://cstr.cn/31254.49.DELTASCIENCE.V1.QI.0007}{\texttt{V1.QI.0007}} & Physics, Engineering & Quantum Comp., Eng. Design & Open Design \& Search & 10h & DP Technology \\
32 & \texttt{QI-0008} & \href{https://cstr.cn/31254.49.DELTASCIENCE.V1.QI.0008}{\texttt{V1.QI.0008}} & Physics, Engineering & Quantum Comp., Eng. Design & Open Design \& Search & 12h & DP Technology \\
33 & \texttt{QI-0011} & \href{https://cstr.cn/31254.49.DELTASCIENCE.V1.QI.0011}{\texttt{V1.QI.0011}} & Physics, Math \& CS & Quantum Comp., Operations Res., Algorithm Design & Open Design \& Search & 4h & DP Technology \\
34 & \texttt{FUS-0001} & \href{https://cstr.cn/31254.49.DELTASCIENCE.V1.FUS.0001}{\texttt{V1.FUS.0001}} & Physics & Fusion \& Plasma & Simulation \& Comp. & 4h & DP Technology \\
35 & \texttt{LKM-0006} & \href{https://cstr.cn/31254.49.DELTASCIENCE.V1.LKM.0006}{\texttt{V1.LKM.0006}} & Chemistry & Comp. Chem., Mol. Spectroscopy & Simulation \& Comp. & 8h & DP Technology \\
36 & \texttt{LKM-0014} & \href{https://cstr.cn/31254.49.DELTASCIENCE.V1.LKM.0014}{\texttt{V1.LKM.0014}} & Engineering & Fluid Mech. & Simulation \& Comp. & 8h & DP Technology \\
37 & \texttt{LKM-0027} & \href{https://cstr.cn/31254.49.DELTASCIENCE.V1.LKM.0027}{\texttt{V1.LKM.0027}} & Chemistry & Comp. Chem., Mol. Spectroscopy & Simulation \& Comp. & 8h & DP Technology, Peking University \\
38 & \texttt{LKM-0048} & \href{https://cstr.cn/31254.49.DELTASCIENCE.V1.LKM.0048}{\texttt{V1.LKM.0048}} & Physics, Materials Science & Comp. Materials, Surface \& Nanomat., Condensed Matter & Simulation \& Comp. & 8h & DP Technology \\
39 & \texttt{LKM-0062} & \href{https://cstr.cn/31254.49.DELTASCIENCE.V1.LKM.0062}{\texttt{V1.LKM.0062}} & Engineering & Fluid Mech., Heat Transfer & Simulation \& Comp. & 8h & DP Technology \\
40 & \texttt{Science-043} & \href{https://cstr.cn/31254.49.DELTASCIENCE.V1.SCIENCE.043}{\texttt{V1.SCIENCE.043}} & Earth Science & Geophysics & Simulation \& Comp. & 5h & DP Technology \\
41 & \texttt{Science-023} & \href{https://cstr.cn/31254.49.DELTASCIENCE.V1.SCIENCE.023}{\texttt{V1.SCIENCE.023}} & Math \& CS & Comp. Math, Machine Learning & Simulation \& Comp. & 6.7h & DP Technology \\
42 & \texttt{Science-017} & \href{https://cstr.cn/31254.49.DELTASCIENCE.V1.SCIENCE.017}{\texttt{V1.SCIENCE.017}} & Physics, Materials Science & Comp. Materials, Condensed Matter & Simulation \& Comp. & 8h & DP Technology, AI for Science Institute, Beijing \\
43 & \texttt{Science-053} & \href{https://cstr.cn/31254.49.DELTASCIENCE.V1.SCIENCE.053}{\texttt{V1.SCIENCE.053}} & Physics, Materials Science & Comp. Materials, Condensed Matter & Simulation \& Comp. & 8h & DP Technology, AI for Science Institute, Beijing \\
44 & \texttt{T8} & \href{https://cstr.cn/31254.49.DELTASCIENCE.V1.T8}{\texttt{V1.T8}} & Physics, Materials Science & Surface \& Nanomat., Stat. Physics & Simulation \& Comp. & 8h & DP Technology \\
45 & \texttt{T14} & \href{https://cstr.cn/31254.49.DELTASCIENCE.V1.T14}{\texttt{V1.T14}} & Materials Science, Math \& CS & Materials Design, Operations Res. & Open Design \& Search & 1h & DP Technology \\
46 & \texttt{CWQ-3} & \href{https://cstr.cn/31254.49.DELTASCIENCE.V1.CWQ.3}{\texttt{V1.CWQ.3}} & Physics, Materials Science & Mech. of Materials, Condensed Matter & Simulation \& Comp. & 16h & DP Technology \\
47 & \texttt{CWQ-4} & \href{https://cstr.cn/31254.49.DELTASCIENCE.V1.CWQ.4}{\texttt{V1.CWQ.4}} & Physics, Materials Science & Stat. Physics, Comp. Materials & Simulation \& Comp. & 48h & DP Technology \\
48 & \texttt{CSF-1} & \href{https://cstr.cn/31254.49.DELTASCIENCE.V1.CSF.1}{\texttt{V1.CSF.1}} & Chemistry & Comp. Chem. & Simulation \& Comp. & 8h & DP Technology, Zhejiang University, AI for Science Institute, Beijing \\
49 & \texttt{CSF-2} & \href{https://cstr.cn/31254.49.DELTASCIENCE.V1.CSF.2}{\texttt{V1.CSF.2}} & Chemistry, Math \& CS & Organic Chem., Catalysis \& Electrochem., Machine Learning & Open Design \& Search & 8h & DP Technology, Zhejiang University, AI for Science Institute, Beijing \\
50 & \texttt{CSF-3} & \href{https://cstr.cn/31254.49.DELTASCIENCE.V1.CSF.3}{\texttt{V1.CSF.3}} & Chemistry & Organic Chem., Comp. Chem. & Multi-Stage Analysis & 12h & DP Technology, Zhejiang University, AI for Science Institute, Beijing \\
51 & \texttt{B01} & \href{https://cstr.cn/31254.49.DELTASCIENCE.V1.B01}{\texttt{V1.B01}} & Physics, Engineering & Fusion \& Plasma, Eng. Design & Open Design \& Search & 8h & DP Technology, Tongji University, Zhejiang University \\
52 & \texttt{LH-1} & \href{https://cstr.cn/31254.49.DELTASCIENCE.V1.LH.1}{\texttt{V1.LH.1}} & Physics & Fusion \& Plasma & Open Design \& Search & 24h & DP Technology, Tongji University, Zhejiang University \\
53 & \texttt{LH-2} & \href{https://cstr.cn/31254.49.DELTASCIENCE.V1.LH.2}{\texttt{V1.LH.2}} & Physics & Fusion \& Plasma & Open Design \& Search & 24h & DP Technology, Tongji University, Zhejiang University \\
54 & \texttt{LH-3} & \href{https://cstr.cn/31254.49.DELTASCIENCE.V1.LH.3}{\texttt{V1.LH.3}} & Physics & Fusion \& Plasma & Open Design \& Search & 24h & DP Technology, Tongji University, Zhejiang University \\
55 & \texttt{M-01} & \href{https://cstr.cn/31254.49.DELTASCIENCE.V1.M.01}{\texttt{V1.M.01}} & Physics, Materials Science & Comp. Materials, Condensed Matter & Simulation \& Comp. & 6h & AI for Science Institute, Beijing, Zhejiang University \\
56 & \texttt{z-01} & \href{https://cstr.cn/31254.49.DELTASCIENCE.V1.Z.01}{\texttt{V1.Z.01}} & Life Sciences, Math \& CS & Bioinformatics, Algorithm Design & Open Design \& Search & 8h & AI for Science Institute, Beijing, Zhejiang University \\
57 & \texttt{z-02} & \href{https://cstr.cn/31254.49.DELTASCIENCE.V1.Z.02}{\texttt{V1.Z.02}} & Math \& CS & Algebra \& Comb., Operations Res. & Open Design \& Search & 8h & AI for Science Institute, Beijing, Zhejiang University \\
58 & \texttt{z-03} & \href{https://cstr.cn/31254.49.DELTASCIENCE.V1.Z.03}{\texttt{V1.Z.03}} & Physics, Materials Science & AMO Physics, Surface \& Nanomat. & Open Design \& Search & 8h & AI for Science Institute, Beijing, Zhejiang University \\
59 & \texttt{q01} & \href{https://cstr.cn/31254.49.DELTASCIENCE.V1.Q01}{\texttt{V1.Q01}} & Life Sciences & Bioinformatics, Omics & Multi-Stage Analysis & 2h & DP Technology \\
60 & \texttt{q03} & \href{https://cstr.cn/31254.49.DELTASCIENCE.V1.Q03}{\texttt{V1.Q03}} & Life Sciences & Bioinformatics, Omics & Multi-Stage Analysis & 2h & DP Technology \\
61 & \texttt{q04} & \href{https://cstr.cn/31254.49.DELTASCIENCE.V1.Q04}{\texttt{V1.Q04}} & Life Sciences & Bioinformatics, Omics & Multi-Stage Analysis & 2h & DP Technology \\
62 & \texttt{q06} & \href{https://cstr.cn/31254.49.DELTASCIENCE.V1.Q06}{\texttt{V1.Q06}} & Life Sciences & Bioinformatics, Omics & Multi-Stage Analysis & 2h & DP Technology \\
63 & \texttt{q07} & \href{https://cstr.cn/31254.49.DELTASCIENCE.V1.Q07}{\texttt{V1.Q07}} & Life Sciences, Physics & Structural Bio., Biophysics & Multi-Stage Analysis & 2h & DP Technology \\
64 & \texttt{q09} & \href{https://cstr.cn/31254.49.DELTASCIENCE.V1.Q09}{\texttt{V1.Q09}} & Chemistry, Life Sciences & Analytical Chem., Omics & Multi-Stage Analysis & 2h & DP Technology \\
65 & \texttt{q10} & \href{https://cstr.cn/31254.49.DELTASCIENCE.V1.Q10}{\texttt{V1.Q10}} & Chemistry, Math \& CS & Cheminformatics, Machine Learning & Multi-Stage Analysis & 2h & DP Technology \\
\end{longtable}
\endgroup

\end{document}